\documentclass[
    pre,
    twocolumn,
    twoside,
    byrevtex,
    superscriptaddress,
    floatfix,
    nofootinbib,
    longbibliography
]{revtex4-1}

\usepackage[realmainfile]{currfile}
\input{\currfilebase.settings}
\usepackage{color}

\usepackage{graphicx,epsfig,verbatim,enumerate}
\usepackage{amssymb,amsmath}
\usepackage{ifthen}

\usepackage{mathtools}

\usepackage{tabularx}
\usepackage{multirow, makecell}
\newcolumntype{C}{>{\centering\arraybackslash}X}
\newcolumntype{L}{>{\raggedright\arraybackslash}X}
\newcolumntype{R}{>{\raggedleft\arraybackslash}X}
\usepackage[latin1,utf8]{inputenc}
\usepackage[normalem]{ulem}
\usepackage{natbib}

\usepackage{subfigure}
\usepackage{longtable}
\usepackage{gensymb}

\usepackage{placeins}

\newboolean{twocolswitch}

\newcommand{\sindex}[1]{}
\newcommand{\nindex}[1]{}

\newcommand{\www}[1]{\url{#1}}

\usepackage{lettrine}

\usepackage[dvipsnames]{xcolor}

\setboolean{twocolswitch}{true}

\begin{document}

\title{\protect
Sentiment and structure in word co-occurrence networks on Twitter
}

\author{
\firstname{Mikaela Irene}
\surname{Fudolig}
}
\email{mikaela.fudolig@uvm.edu}

\affiliation{
  Computational Story Lab,
  Vermont Complex Systems Center,
  MassMutual Center of Excellence for Complex Systems and Data Science,
  Vermont Advanced Computing Core,
  University of Vermont,
  Burlington, VT, USA
  }

\author{
\firstname{Thayer}
\surname{Alshaabi}
}

 \affiliation{
  Advanced Bioimaging Center,
  UC Berkeley,
  Berkeley, CA, USA
  }
  
\affiliation{
  Computational Story Lab,
  Vermont Complex Systems Center,
  MassMutual Center of Excellence for Complex Systems and Data Science,
  Vermont Advanced Computing Core,
  University of Vermont,
  Burlington, VT, USA
  }

\author{
\firstname{Michael V.}
\surname{Arnold}
}

\affiliation{
  Computational Story Lab,
  Vermont Complex Systems Center,
  MassMutual Center of Excellence for Complex Systems and Data Science,
  Vermont Advanced Computing Core,
  University of Vermont,
  Burlington, VT, USA
  }

\author{
\firstname{Christopher M.}
\surname{Danforth}
}

\affiliation{
  Computational Story Lab,
  Vermont Complex Systems Center,
  MassMutual Center of Excellence for Complex Systems and Data Science,
  Vermont Advanced Computing Core,
  University of Vermont,
  Burlington, VT, USA
  }

\affiliation{
  Department of Mathematics \& Statistics,
  University of Vermont,
  Burlington, VT, USA
  }

\author{
\firstname{Peter Sheridan}
\surname{Dodds}
}

\affiliation{
  Computational Story Lab,
  Vermont Complex Systems Center,
  MassMutual Center of Excellence for Complex Systems and Data Science,
  Vermont Advanced Computing Core,
  University of Vermont,
  Burlington, VT, USA
  }

\affiliation{
  Department of Computer Science,
  University of Vermont,
  Burlington, VT, USA
}

\date{\today}

\begin{abstract}
  \protect
  We explore the relationship between context and happiness scores in political tweets using word co-occurrence networks, where nodes in the network are the words, and the weight of an edge is the number of tweets in the corpus for which the two connected words co-occur. In particular, we consider tweets with hashtags \#imwithher 
and 
\#crookedhillary, 
both relating to Hillary Clinton's presidential bid in 2016.
We then analyze the network properties in conjunction with the word scores by comparing with null models to separate the effects of the network structure and the score distribution. 
Neutral words are found to be dominant and most words, regardless of polarity, tend to co-occur with neutral words. 
We do not observe any score homophily among positive and negative words. 
However, when we perform network backboning, 
community detection results in word groupings with meaningful narratives, 
and the happiness scores of the words in each group correspond to its respective theme. 
Thus, although we observe no clear relationship between happiness scores and co-occurrence at the node or edge level, 
a community-centric approach can isolate themes of competing sentiments in a corpus. 
\end{abstract}



\maketitle


\section{Introduction}
\label{sec:introduction}

 Large-scale analysis of user-generated text has been instrumental in understanding recent political campaigns and movements, from the Arab Spring~\cite{howard_opening_2011,wolfsfeld_social_2013} to the Black Lives Matter protests~\cite{wu_say_2021}.
Complementing traditional survey based approaches, social media has also shown promise as a medium for gauging public sentiment.
Unlike surveys, however, where questions are carefully constructed so that the answers can be directly interpreted, extracting sentiment from large-scale unstructured text requires automated interpretation. The field of sentiment analysis in natural language processing (NLP) developed in response to this need, not just in politics but in other areas as well. Retailers have used sentiment analysis techniques to analyze product reviews and provide a better experience for its users~\cite{fang_sentiment_2015,shivaprasad_sentiment_2017}. Sentiment analysis has also been applied to financial markets, where the opinion of market participants plays an important role in future prices~\cite{smailovic_predictive_2013,pagolu_sentiment_2016,mishev_evaluation_2020}. Understanding public opinion also matters in public health, as in the case of addressing perceptions of vaccines~\cite{raghupathi_studying_2020,klimiuk_vaccine_2021}, and predicting signals of depression in self-expressed social media posts~\cite{wang_depression_2013,coppersmith_quantifying_2014,reece_forecasting_2017,stupinski_quantifying_2021}.

Studies on sentiment in social media can be broadly divided into two groups: one where the emphasis is on identifying the sentiment of a particular statement, and another that focuses on obtaining a collective measure of sentiment. In this work, we deal with the latter, where we are interested in the general sentiment associated with a particular topic rather than the emotional state expressed by a particular individual. This paradigm of studying collective sentiment has been used to monitor general sentiment on Twitter, and has been validated by cross-referencing changes in measured sentiment with high-profile events, such as election results, terrorist activities, holidays and even birthdays of popular artists~\cite{dodds_temporal_2011}. For example, holidays such as Christmas, New Year, Fourth of July and Thanksgiving correspond to spikes in happiness, while mass shootings in the United States, the fire at Notre Dame cathedral, the murder of George Floyd and the storming of the US Capitol coincide with a marked decrease in happiness on Twitter.
In contrast to this approach, which looks at social media posts as a bag of words, we look at it as a collection of tweets that are in turn related by their use of specific words. In particular, do certain words go together, and if so, what is the sentiment profile of these groups of words, and what do they say about the sentiment profile of the tweets themselves? We want to know if this approach will provide a more nuanced picture of collective sentiment that is missed by a simple aggregation.
 
One way to infer collective sentiment or meaning from text is through lexicons, where a subset of words is compiled, either through expert curation or by frequency of use, to provide a representative sample of the corpus. 
Sentiment scores are then assigned to each word, usually by human annotators. One such lexicon is the labMT dataset used in the Hedonometer~\cite{dodds_temporal_2011,dodds_human_2015}, which contains happiness scores from human annotators for more than 10,000 words. The word list was later expanded~\cite{alshaabi_augmenting_2021} to extrapolate happiness scores for any unscored word using a word embedding model. Aside from happiness scores, semantic differentials have also been used to quantitatively capture meanings of words. The valence-arousal-dominance (VAD) semantic differentials have been commonly used since its inception, and a number of studies have focused on creating large-scale VAD lexicons~\cite{bradley_affective_1999,warriner_norms_2013,mohammad_obtaining_2018}. Other examples of sentiment lexicons include SentiWordNet~\cite{baccianella_sentiwordnet_2010} and LIWC~\cite{pennebaker_development_2015}.

In contrast to lexicon scores, co-occurrence models infer the meaning of a word using the surrounding context, i.e., the adjacent words that occur proximate to the anchor word in real-world corpora. These models output word embeddings, which are high-dimensional vectors that represent a word's meaning. Examples of such models are Word2Vec~\cite{mikolov_distributed_2013}, GloVe~\cite{pennington_glove_2014}, and FastText~\cite{bojanowski_enriching_2017,joulin_bag_2017}. Although these word embeddings perform well in complex tasks such as machine translation and language detection, the meaning is abstracted in several dimensions and is thus less interpretable than the scores provided by lexicons. A study by \citet{hollis_principals_2016} examined whether the 300 dimensions of Word2Vec corresponds to any of the well-known semantic and lexical variables, including valence, arousal and dominance. However, they found that no single dimension corresponded to a particular semantic or lexical variable, although the principal components that contributed the most to the variance were found to be correlated to some semantic and lexical variables. This is particularly interesting as Word2Vec was not developed using these semantic and lexical concepts, but nevertheless captures aspects of them in the embedding.

Conversely, context is not used to obtain lexicon scores; human annotators are presented with words and are asked to score them in the desired scale. The interplay between sentiment scores and co-occurrence patterns in words has mostly been viewed from a predictive standpoint, i.e., whether sentiment scores can be predicted using word co-occurrence. Early work predicted sentiment polarity (positive or negative)~\cite{turney_unsupervised_2002,turney_measuring_2003}, while later models predicted valence, arousal, and dominance using latent semantic analysis~\cite{bestgen_checking_2012}, pointwise mutual information~\cite{recchia_reproducing_2015} and co-occurrence models such as HiDEx~\cite{shaoul_word_2006,westbury_avoid_2015} and Word2Vec~\cite{hollis_extrapolating_2017}. Recently, Alshaabi et al~\cite{alshaabi_augmenting_2021} used FastText word embeddings and the transformer model DistilBERT~\cite{sanh_distilbert_2019}, both pre-trained on large-scale, general corpora, to predict happiness scores for out-of-vocabulary words. While prediction success implies the existence of a relationship between word co-occurrence and lexicon scores, we are not aware of any study that delves deeper into the details of this relationship. In particular, given a corpus, how are the sentiment scores of co-occurring words related? And can the co-occurrence structure of words be used to uncover mixed sentiments in a corpus that would be missed by an aggregate measure, such as the average happiness score?

We explore these questions in the context of tweets, which because of character count limitations (140 before 2018, and 280 thereafter) generally contain words that are related by a single theme. Rather than using techniques resulting in word embeddings that abstract words into non-interpretable vectors, we retain the nature of the word itself and instead use tools from network science to analyze the word co-occurrence structure in the corpus. Although NLP has leaned more towards probabilistic models and neural networks, network science has nevertheless found a wide range of applications in computational linguistics~\cite{ferrer_i_cancho_patterns_2004,liu_language_2013,cong_approaching_2014,al_rozz_characterization_2017,wang_learning_2017,chen_how_2018,jiang_does_2019}. In fact, some techniques commonly used in NLP, such as latent Dirichlet allocation, have been shown to parallel network science techniques~\cite{gerlach_network_2018}. By constructing word co-occurrence networks, where the words are nodes in a network, the sentiment scores are node properties, and the edges in the network represent the co-occurrence structure, we can see more directly how co-occurrence and sentiment are related.

Specifically, each node in the network represents a word found in a tweet. As a tweet contains words that are related to each other, each tweet corresponds to a connected subgraph, where each word appearing in the tweet is connected to every other word appearing in the same tweet. We superpose the subgraphs for each tweet in the corpus to form the full word co-occurrence network, with the edge weights corresponding to the number of tweets in which two words co-occur together. The degree and strength of a node tell us how often a word co-occurs with others, while the weight of an edge indicates how often any two words co-occur.

We then analyze the context-dependent network structure and its relation to the context-independent word lexicon scores, in particular the happiness scores from both the labMT dataset and the predictive model in~\cite{alshaabi_augmenting_2021}. We look at the network from three perspectives: node-centric, edge-centric, and community-centric. In the node-centric perspective, we examine if there is a relationship between node characteristics, such as degree and node strength, and the happiness scores. In the edge-centric perspective, we consider pairs of words and their respective scores. Lastly, in the community-centric approach, we use network backboning and community detection techniques to obtain groups of related words within the network, and explore the relationship of scores within each group to scores across groups. 

As we are interested in whether we can uncover mixed sentiments in corpora, we focus on political tweets, where polarization in stance is common. Stance is distinct from sentiment, in that it reflects the opinion (``favor'', ``against'', or ``neither'') towards a target~\cite{kucuk_stance_2020}. Regardless, sentiment has been shown to be important, though not sufficient, to predict stance~\cite{mohammad_stance_2016}. Thus, by combining tweets using hashtags that are from opposite stances, we are likely to get tweets with opposing sentiments. Further, within each stance, we will also likely encounter a minority of tweets with a sentiment opposite to that expected, e.g., an ``against'' stance tweet may have a positive sentiment towards the target's opponent. We also examine whether these cases will be picked up by our analysis. As the political situations associated with these tweets are likely to be reported in the news, we can use domain knowledge to check the validity of our results.

We detail the selection of tweets and the method of our analysis in Section~\ref{sec:method}. In Section~\ref{sec:results}, we present our findings regarding the relationship between the co-occurrence network structure and happiness scores at the node-centric, edge-centric, and community-centric levels. We summarize our results in Section~\ref{sec:conclusion} and present avenues for future work.

\section{Methodology}
\label{sec:method}

\subsection{The tweets}
\label{subsec:method-tweets}

For our analysis, we consider a collection of tweets relating to Hillary Clinton's presidential bid, in particular, tweets with any of the following anchors, \texttt{\#imwithher} or \texttt{\#crookedhillary} (case-insensitive). Both anchors were mostly used in 2016 during the US presidential election season, where \texttt{\#imwithher} is associated with being in favor of Hillary Clinton and \texttt{\#crookedhillary} associated with being against her. Using Twitter's Decahose API, we take a random 10\% of the tweets matching these hashtags during the 1-week period of election season when the anchors were most popular, respectively. These periods were November 1-8 for \texttt{\#imwithher} and October 7-14 for \texttt{\#crookedhillary}, both in 2016~\cite{alshaabi_storywrangler_2021}. We acknowledge that although these stance association assumptions are general, they are not universal, and each hashtag may be used for the opposite stance. We also investigate whether such cases can be uncovered using our analysis.

We then sampled the tweets so that we get balanced classes, with each hashtag getting 7222 unique original tweets (no retweets, quote tweets, or duplicates). We also note that we found an obvious case of systematic hashtag hijacking with content unrelated to Hillary Clinton and removed the corresponding tweets before sampling (see the Supplementary Information for details).  The tweets containing \texttt{\#imwithher} are labeled as ``favor'' tweets, while the ones containing \texttt{\#crookedhillary} are labeled as ``against'' tweets. These labels correspond to the general stance associated with each anchor hashtag. The aggregated set of tweets is denoted by the label ``all''.

\subsection{Happiness scores}
\label{subsec:method-scores}

For the word ratings, we consider the Language Assessment by Mechanical Turk (labMT) dataset that contains scores for happiness~\cite{dodds_temporal_2011,dodds_human_2015} for more than 10,000 words. The list was compiled from the most frequently appearing words in more contemporary sources such as Twitter, Google Books, the New York Times and music lyrics. Human annotators were asked to score a word using a scale of 1--9, with 1 being the least happy and 9 being the happiest, and 5 being neutral. The means of the measurements as well as their standard deviations are reported in the dataset.

To get the happiness scores for words not present in the labMT dataset, we use deep learning models trained on the labMT dataset to extrapolate happiness scores for words and n-grams~\cite{alshaabi_augmenting_2021}. The first proposed model, namely the token model, breaks up the input into character-level $n$-grams and uses subword embeddings to estimate their happiness, while the second model uses dictionary definitions to gauge happiness scores. The predictive scoring models have been found to be reliable, making predictions well within the range of scores that are obtained from different human annotators. Although the dictionary model provides better performance for words in the dictionary, the token model performs almost as well with lower computation cost and gives more reliable results for words that are not in dictionaries but are made up of component subwords, such as hashtags. We use the token model to infer the happiness scores for the words in the network that are not in the original labMT dataset, allowing us to compare the effect of the choice of words to evaluate the network structure. To further improve accuracy, we use the expanded form of the most commonly used acronyms in the corpora examined (Table~\ref{tab:acronyms}) as an input to the predictive scoring model to obtain their respective scores.

\begin{table}
\caption{\textbf{Acronyms and their full versions.} Acronyms in the corpus with high word counts are scored using their full versions in the predictive scoring model using tokens~\cite{alshaabi_augmenting_2021}. The model breaks the input token into character-level $n$-grams, so there is no need to add whitespaces here.}
\label{tab:acronyms}
{%
\begin{tabular}{ll}
\textit{Acronym in tweet} & \textit{Model input} \\ \hline
maga                      & makeamericagreatagain       \\
msm                       & mainstreammedia             \\
tcot                      & topconservativesontwitter  
            \\
potus                      & presidentoftheunitedstates 
\end{tabular}%
}
\end{table}

\subsection{Generating the co-occurrence network}
\label{subsec:method-network_build}

We parse each tweet as follows. As the writing style on Twitter involves a liberal use of hashtags, we disregard the hashtag symbol in each tweet. Each tweet is then converted into lower case and parsed by splitting each string into a set of words separated by whitespaces, while preserving contractions, punctuation, handles, hashtags, dates and links (see~\cite{alshaabi_storywrangler_2021} for more information).
For simplicity, we further exclude punctuation marks, and undo contractions, as shown in Table~\ref{tab:contractions}. Twitter handles (1-grams beginning with a ``@''), 1-grams with numbers, and URL links were also removed, as well as the anchor terms \texttt{\#crookedhillary} and \texttt{\#imwithher} in order to eliminate community structure associated with the choice of these hashtags. In our preliminary analysis, we find that the names of Hillary Clinton and her opponent, Donald Trump, also effectively served as anchor words, masking finer structure in the network. We thus also remove the terms ``hillary'', ``clinton'', ``hillaryclinton'', ``trump'', ``donald'', and ``donaldtrump''. 

We then obtain counts for the words used in the tweets (the document-term matrix) using the Python package \texttt{scikit-learn}~\cite{pedregosa_scikit-learn_2011}. 
In case a word appears multiple times inside a single tweet, it is only counted once in generating the network; however, the original word counts are also recorded. We finally obtain the adjacency matrix that is used to generate the undirected word co-occurrence network by taking the dot product of the document-term matrix (where each word is only counted once in a tweet) and its transpose, setting the diagonal elements to zero.
In the resulting network, each node is a word occurring in the set of tweets. An edge between two nodes indicates that the two nodes occur together in some tweets, and the edge weights tells us how many tweets the two connected nodes occur together. We construct a separate network for each set of tweets (e.g., for the Hillary Clinton tweet corpus, we create a separate network for the ``favor'', ``against'', and ``all'' tweets).

\begin{table*}[]
\caption{\textbf{Contractions and equivalents.} Some contractions in tweets were converted to equivalent strings. In case of ambiguity, the contraction is converted to a blank space.}
\label{tab:contractions}
{%
\begin{tabular}{lll}
\textit{Contractions} & \textit{Converted to}              & \textit{Remarks}                                    \\ \hline
n't                   & \textless{}space\textgreater{}not  &                                                     \\
's                    & \textless{}space\textgreater{}     & either ``is" or possessive marker                    \\
'm                    & \textless{}space\textgreater{}     & ``am"; for consistency, since ``'s" is converted to a space \\
'd                    & \textless{}space\textgreater{}     & either ``had" or ``would"                             \\
're                   & \textless{}space\textgreater{}     & ``are"; for consistency, since ``'s" is converted to a space \\
've                   & \textless{}space\textgreater{}have &                                                     \\
'll                   & \textless{}space\textgreater{}will &                                                    
\end{tabular}%
}
\end{table*}

\subsection{Network characterization}
\label{subsec:method-network-characterization}

We look at the basic properties of the network, such as the node strength, degree and edge weight distribution, and examine how these are related to the scores of the node. To determine if the results we observe are due to the happiness scores or to the network structure, we compare our results with a number of null models, where we modify either the network structure or the scores of the words.  
These include:

\paragraph{Configuration model}
The network is randomly rewired using the node strength sequence to roughly preserve the node strength distribution; self-loops are discarded, and parallel unweighted edges connecting two nodes are combined to form a weighted edge~\cite{serrano_weighted_2005}
     
\paragraph{Erdos-Renyi model}
An Erdos-Renyi network is constructed with the probability $p$ of an edge occurring set to be $E_\mathrm{obs} / E_\mathrm{max}$, where $E_\mathrm{obs}$ is the number of observed (unweighted) edges and $E_\mathrm{max}$ is the total number of possible (unweighted) edges. The edge weights are assigned by drawing with replacement from the distribution of edge weights in the original graph.

\paragraph{Shuffled score model}
The network is kept as is, but the scores are reshuffled among the nodes.

\paragraph{Uniform score model}
The network is kept as is, but the scores of the nodes are sampled from a uniform distribution from 1 to 9 (the prescribed range of happiness scores in labMT).

\subsection{Community detection and network backboning}
\label{subsec:method-community}

To explore whether clusters in the co-occurrence network are also related to sentiment scores, we perform community detection on the co-occurrence network. However, because the resulting co-occurrence networks have words that occur very frequently (such as the words ``is'' and ``the'') and also many words that are used together in only a few tweets, we find that performing community detection on the full set of words resulted in communities that were dominated by function words and did not correspond to any discernible theme. We thus need to trim down the network to expose a more robust underlying structure. To do this, we use \textit{network backboning} techniques, which remove edges that are likely to occur at random. In particular, we apply the disparity filter~\cite{serrano_extracting_2009} and the noise-corrected model~\cite{coscia_network_2017} using the implementation given by \citet{coscia_network_2017}. Compared to other recent network backboning methods~\cite{grady_robust_2012,slater_two-stage_2009}, the disparity filter and the noise-corrected model allow for tunable parameters and provide a theoretical foundation to address noisy edge weights. The disparity filter looks at each node and the normalized weights of the $k$ edges connecting it to its neighbors. These weights are compared to a null model, where the interval $[0,1]$ is subdivided by $k-1$ points uniformly distributed along it. The lengths of each segment would represent the expected values for the $k$ normalized weights, and the probability of each edge having a weight compatible with this null model is calculated, corresponding to a p-value. Those edges compatible with the null model at a preset statistical significance level $\alpha$ are removed from the backbone. The noise-corrected model is similar, but looks at a node pair rather than a node in constructing a null model. In the noise-corrected model, the edge weight distribution is assumed to be binomial, and a preset threshold analogous to the p-value decides which edges to keep in the backbone.

In most real-world networks, hubs are considered crucial to the structure and dynamics; transportation and friendship networks are notable examples where this holds. Similarly, edges that have high weights are also considered important and will not be eliminated by network backboning algorithms. However, in our word co-occurrence networks, this is not necessarily so. For example, consider the words ``is'' and ``the'', two of the network nodes with the highest degrees, which are also connected by an edge with one of the highest weights. As both are important components in constructing English sentences, this prominence is not unexpected. However, in terms of characterizing the meaning of a corpus, neither of these words is useful. Backboning of the word co-occurrence network requires removing these hubs as well. A tricky complication is that not all hubs are function words. For example, for the Hillary Clinton corpus, there are also words such as ``vote'' or ``maga'' that are relevant to the meaning of the corpus, but are similar to the function words ``is'' and ``the'' in that they are high-degree nodes that are connected by a high-weight edge.

To differentiate between these two types of high-degree nodes, we create a list of words that are frequently used among several corpora. 
One option is to use the list of stop words included in the Python natural language toolkit (NLTK)~\cite{loper_nltk_2002}, which includes pronouns, prepositions, articles, and other function words. This gives us 180 words (including the preposition ``us'' which we added, as we saw it remained a prominent pronoun in the corpus after our analysis), some of which include the list of contractions we removed in parsing. Another option is to use the most commonly seen words on Twitter, which are less likely to have relevant meaning in reference to the anchored set.

Using Storywrangler~\cite{alshaabi_storywrangler_2021}, we identify the top 400 case-sensitive 1-grams for 100 randomly chosen days from 2010/01/01 to 2015/12/31 (prior to the period corresponding to our corpus tweets) and take the intersection of these 100 lists. The resulting words frequently occur on Twitter, indicating that they are not likely to hold much distinctive meaning. After removing punctuation marks, digits and symbols, and case-insensitive duplicates from this list, we get 194 commonly used words on Twitter. We then take the intersection of this list with the top 200 words in the network with the highest degrees. The final list, which depends on the corpus, is the resulting list of stop words. Both the NLTK and the top Twitter 1-gram method give similar results. For brevity, we present the results obtained using the Twitter 1-gram method. A table containing the words removed using this method is included in the Supplementary Information.

We perform network backboning in two passes. We first eliminate the words that are likely to hold less distinctive meaning using the top ranked words on general Twitter. This is equivalent to disregarding stop words in parsing, where the stop words are based on general Twitter and are only removed if they serve as hubs in the network. We then remove irrelevant edges using the disparity filter or the noise-corrected model. With these steps, both the most irrelevant and the most connected words are removed to obtain the network backbone, which is in turn used for further analysis.

We then perform community detection on the network backbone. Although an overlapping community detection algorithm is the natural choice for this task, we found that using hierarchical link clustering~\cite{ahn_link_2010} gives clusters with overlapping themes. We present instead the results obtained using the Louvain algorithm~\cite{blondel_fast_2008}, a widely used greedy optimization method that assigns each node to a community and returns the partitioning that maximizes the graph's modularity. Surprisingly, despite the limitation that the communities must be disjoint, it gives the word co-occurrence network an interpretable community structure. Each community is then analyzed in terms of the words it contains and the corresponding word scores. With a non-overlapping community detection algorithm, since a word only appears once in every community, the score contributions of a particular word to the community and to the unpartitioned network can also be easily interpreted.

\section{results}
\label{sec:results}

We begin by presenting the basic characteristics of the corpora, such as the degree, count, and score distributions. We then look at how node-centric properties, such as the node strength and degree, relate with word scores, and how the scores of the words connected by each edge are related. 

\subsection{Network characteristics}

\begin{table*}[!tp]
\caption{\textbf{Number of tweets, nodes, node strengths, degree and the size of connected components for tweets.}}
\centering
\label{tab:anchored_tweets_describe}
{%
\begin{tabularx}{\textwidth}{l CCCCr}
\multicolumn{1}{c}{\textit{Anchors}}                         & \multicolumn{1}{c}{$N_\mathrm{tweets}$} & \multicolumn{1}{c}{$N_\mathrm{nodes}$} &
\multicolumn{1}{c}{\textit{Total edge weight}}      &
\multicolumn{1}{c}{\textit{Number of edges}} & \multicolumn{1}{c}{\textit{Component sizes}} \\ \hline
\#imwithher                     & 7222      & 6085    & 408303 & 186688 & (6080, 1$\times$5) \\
\#crookedhillary                & 7222      & 9912     & 489294 & 264226 & (9898, 3, 3, 1$\times$8) \\
\#imwithher OR \#crookedhillary & 14444     & 12023    & 897597 & 405448 & (12005, 3, 3, 1$\times$12)
\end{tabularx}%
}
\end{table*}

\begin{figure*}[ht!]
    \centering
    \includegraphics[width=\textwidth]{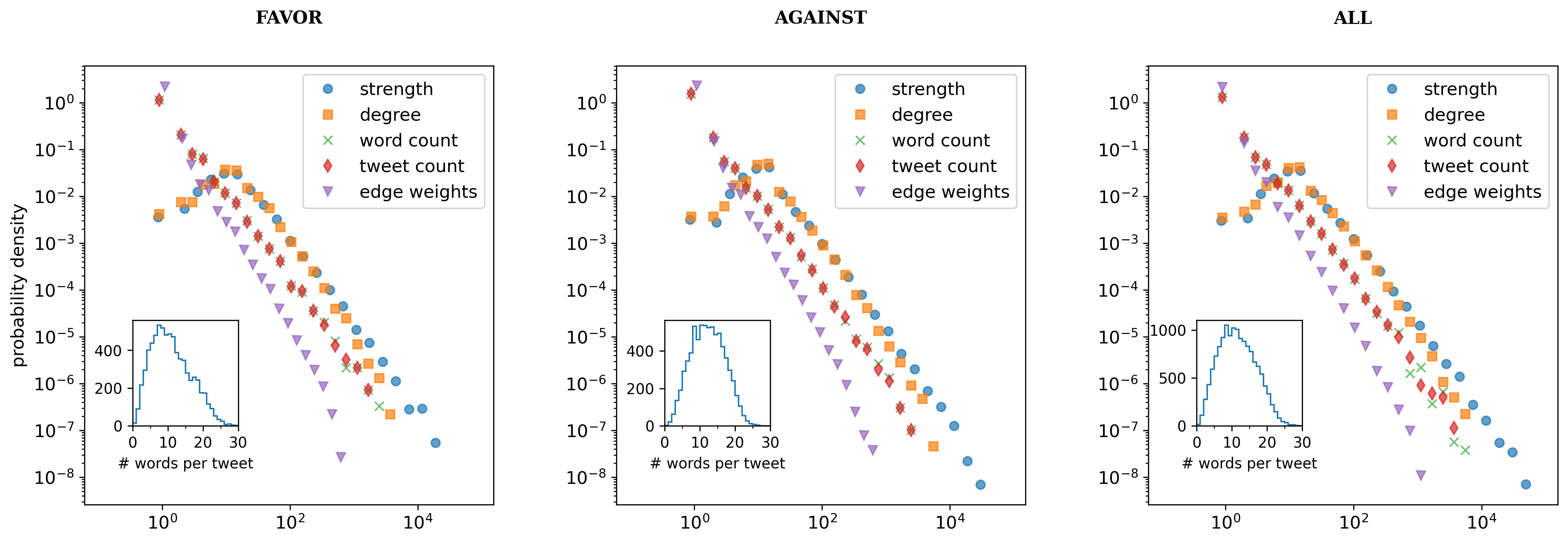}
    \caption{\textbf{Word co-occurrence network statistics.} The plots show the degree, word count, tweet count, and edge weight distribution for the word co-occurrence networks corresponding to the tweets in favor of and against Hillary Clinton, as well as the combination of these two corpora. The inset shows the number of tweets with a given number of words, or equivalently, the histogram of the sizes of subgraphs, with each subgraph corresponding to a single tweet. The peak of this histogram at around 10 words per tweet explains the dip found in the probability distributions for node strength and degree.}
    \label{fig:hclinton_degree_count_dist}
\end{figure*}

The number of nodes, edges, and tweets for each set of anchored tweets are given in Table~\ref{tab:anchored_tweets_describe}. The distributions for the node strengths, node degrees and edge weights, as well as the word counts (the number of times a word appears in the corpus) and the tweet counts (the number of tweets a word is in), are shown in Figure~\ref{fig:hclinton_degree_count_dist}. The tweet and word counts display a heavy-tailed distribution, as expected~\cite{ryland_williams_zipfs_2015,williams_text_2015}, while the tail of the degree and node strength distributions follow a power law with similar exponents. Because the tweets are short and words are not often repeated in each tweet, the construction of the co-occurrence network creates a high correlation between the degree of a node and the corresponding word count. However, even if many words are usually mentioned in very few tweets, they will co-occur with other words and will have degrees higher than 1. This is consistent with the distribution of the number of nodes extracted per tweet (insets of Figure~\ref{fig:hclinton_degree_count_dist}) and explains the dip in the probability density for the degree distribution below the value of 10, while the probability densities for the word and tweet counts are monotonic. We also note that for the vast majority of nodes, the degree is the same as the node strength. The network also has mild degree disassortativity and low node strength disassortativity (Table~\ref{tab:hclinton_assortativity}).

\begin{table*}[]
\caption{Assortativity coefficients}
\centering
\label{tab:hclinton_assortativity}
{%
\begin{tabularx}{\textwidth}{l CCCCC}
                                &
                                &
                        &         \multicolumn{2}{c} {Happiness}                      \\
\multicolumn{1}{c}{\textit{Anchors}}                         &
\multicolumn{1}{c}{Strength} &
\multicolumn{1}{c}{Degree} &
\multicolumn{1}{c}{Weighted} &
\multicolumn{1}{c}{Unweighted} 
\\ \hline
\#imwithher                     & -0.189      & -0.278    & 0.022 & 0.031  \\
\#crookedhillary                & -0.175      & -0.254     & 0.006 & 0.008  \\
\#imwithher OR \#crookedhillary & -0.166     & -0.261    & 0.022 & 0.024 
\end{tabularx}%
}
\end{table*}

\subsection{Happiness scores and word counts}
\label{subsec:happiness-scores-word-counts}

\begin{figure*}[ht!]
    \centering
    \includegraphics[width=\textwidth]{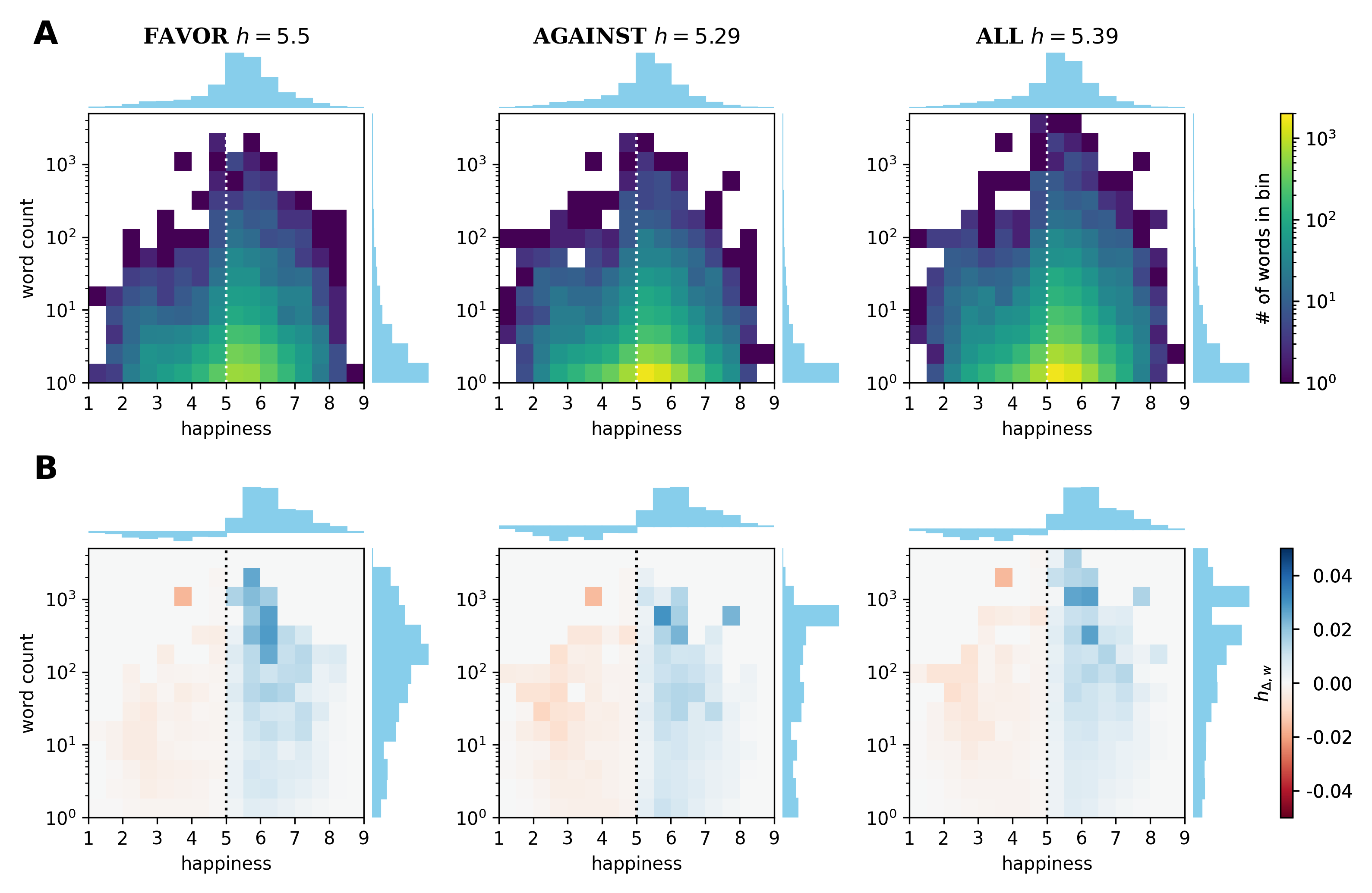}
    \caption{\textbf{Distribution of words and their contributions to the deviation from neutrality in terms of their word counts and happiness scores.} (a) The 2D histogram for word count vs. happiness score, with the corresponding marginal distributions shown (note that each word has a weight of 1 in the marginal distribution). (b) A 2D histogram of the contributions of words in word count-happiness space to the deviation from neutrality, $h_{\Delta, w} = (h_w - 5) * N_w / \sum_{w^{\prime}}{N_{w^{\prime}}}$, where $h_w$ is the word's happiness score of word and $N_w$ is the number of times the word appears in the corpus. The marginal distributions are also included. Vertical lines at $h=5$ are added to guide the eye.
    }
    \label{fig:hclinton_scorecontrib_count_dist_v2}
\end{figure*}

To differentiate the contributions of frequent and rare words as well as positive, neutral and negative words, we construct a 2D histogram for the word counts vs. the happiness scores of the words in each corpus (Figure~\ref{fig:hclinton_scorecontrib_count_dist_v2}a). Words close to the prescribed neutral score $h=5$ are dominant in each corpus, especially among frequently used words. The average happiness score per corpus, computed by weighting the score of each word by its word count, is given in the annotations above the plots. To see which score ranges influence the average score the most, we construct a 2D histogram where each word is weighted by its contribution to the average happiness score. In Figure~\ref{fig:hclinton_scorecontrib_count_dist_v2}b, each word $w$ is given the weight

\begin{equation}
h_{\Delta, w} = (h_w - 5) * N_w / \sum_{w^{\prime}}{N_{w^{\prime}}}
\label{eq:score_contrib}
\end{equation}

\noindent where $h_w$ is the happiness score of word $w$ and $N_w$ is the number of times the word appears in the corpus. Although neutral words are not expected to contribute much to this deviation, there are a few words with happiness scores $h\in(5,6)$ that raise the average happiness significantly, consistent with the observed positivity bias in language~\cite{dodds_human_2015,aithal_positivity_2021}. The average happiness scores in these corpora, however, are lower than those generally obtained with the entire Twitter Decahose dataset, which is also observed in other subsets containing political tweets~\cite{cody_public_2016}. We also note that while low-count words do not individually contribute much to the average happiness score, their ubiquity in the corpus increases their combined influence. As expected, those with higher word counts have higher values of $h_{\Delta, w}$.

\subsection{Network structure and happiness scores}
\label{subsec:results-network_structure_and_happiness}
We now look at how the happiness scores and the network structure are related. We begin by looking at the relationship between a node's degree and node strength to its score, and then that of an edge and the scores of the nodes it connects.

\subsubsection{Node degree, node strength and happiness scores}
\begin{figure*}[ht!]
    \centering
    \includegraphics[width=\textwidth]{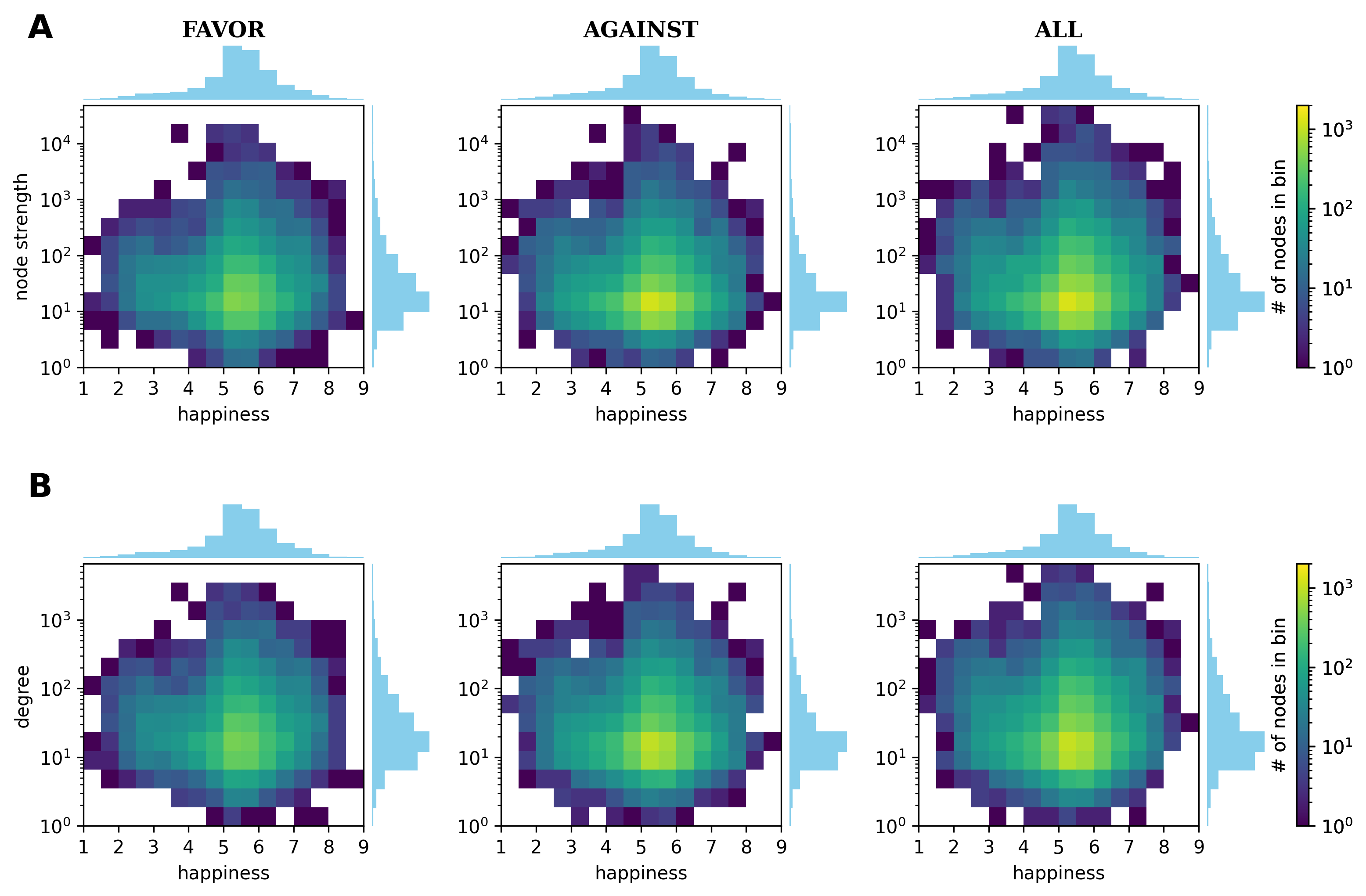}
    \caption{\textbf{Node characterization by node strength, degree, and happiness scores.} 2D histograms for both the (a) node strength and (b) degree vs. happiness score. Note the dominance of nodes in the neutral range, with the happiness score close to $h=5$.}
    \label{fig:hclinton_score_degree_dist_v2}
\end{figure*}

A 2D histogram showing the distribution of words in node strength-score space is given in  Figure~\ref{fig:hclinton_score_degree_dist_v2}. The dominance of neutral words for all values of node strengths and degrees is evident. Comparing these results to the null models (Figures~\ref{fig:hclinton_score_degree_dist_v2_configmodel}-\ref{fig:hclinton_score_degree_dist_v2_uniform}), we find that neither shuffling the scores nor changing the network structure results in a different profile, while keeping the network and using a uniform distribution for the scores changes the result substantially. Thus, this relationship between the node's score and its degree and strength is mostly due to the dominance of neutral words in the score distribution and not the structure of co-occurrence of words in the corpus.

\subsubsection{Happiness scores of connected words}

\begin{figure*}[ht!]
    \centering
    \includegraphics[width=\textwidth]{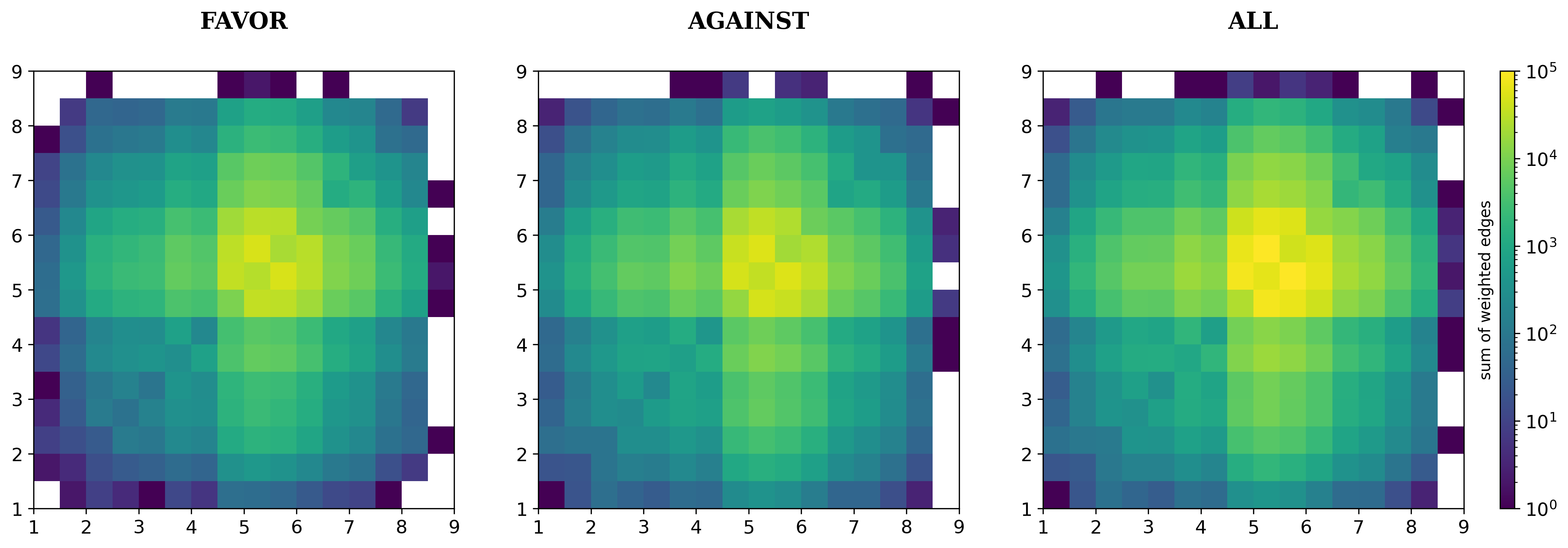}
    \caption{\textbf{Happiness scores for each pair of connected nodes.} Each pair of nodes is weighted by the weight of the edge connecting them. We made the histogram to be symmetric about the 45$\degree$ line so that one can analyze it from either the horizontal or vertical direction.}
    \label{fig:hclinton_scorepair_happiness_heatmap_v2}
\end{figure*}

We now look at the property of each edge, in particular the scores of connected nodes and their edge weights. Figure~\ref{fig:hclinton_scorepair_happiness_heatmap_v2} shows the profile of the scores of connected nodes in the network. We see that words of any score tend to be connected to words with happiness scores $h\in[4.5, 6.5)$. Whereas words in this region tend to co-occur with words of similar scores, words with more extreme scores do not exhibit such homophily. This is consistent with the low assortativity coefficients obtained for the happiness scores (Table~\ref{tab:hclinton_assortativity}) computed using both the weighted and unweighted versions of the network. Comparing these results with null models (Figures~\ref{fig:hclinton_scorepair_happiness_heatmap_v2_configmodel}-\ref{fig:hclinton_scorepair_happiness_heatmap_v2_uniform}), we find that rewiring the network and shuffling the scores give similar profiles, while keeping the network but using a uniform distribution changes the profile drastically. Thus, this homophily among neutral words is mainly due to the abundance of neutral words in the corpus and not the co-occurrence network structure.

\subsection{Network backboning}
\label{subsec:results-network_backboning}

We saw in Section~\ref{subsec:results-network_structure_and_happiness} that the network itself plays a smaller role than the score distribution, both from a node-centric perspective (score, degree, and node strength), and a edge-centric perspective (score difference and edge weights). In particular, the dominance of neutral words in the corpus makes it more likely that any two connected words are close to neutral and thus have similar scores. Previous work has observed this dominance of neutrality in text corpora which tends to reduce the signal in differences in sentiment across time~\cite{dodds_temporal_2011}. This issue was previously addressed by using a manually tuned filter to remove neutral words while still capturing time series correlations~\cite{dodds_temporal_2011}; two commonly used filters exclude words with scores between 4.5 and 5.5 or 4 and 6 before taking the average of the happiness scores (weighted by word count) of the remaining words in the corpus.

Although the removal of neutral words may be sufficient in obtaining an average happiness score for a given corpus, this may not be the best approach when the aim is to explore the relationship between network structure and scores, as removing neutral words in performing network analysis begs the question by presupposing the existence of this relationship. Further, as neutral words form a big chunk of the network regardless of node degree, node strength or edge weight, disregarding them will remove possibly influential nodes that may help us understand hidden substructures in the network, such as opposing sentiments within the corpus, that would not be visible if we look at just the weighted average of the happiness scores. Some context-relevant words (e.g., ``vote'') may also be neutral, and the removal of these words may make it more difficult to interpret the co-occurrence network structure. We want to see if we can obtain useful word sentiment information from the network structure without resorting to score cutoffs.

Network backboning uncovers relevant structure in a network by removing edges that are likely to be spurious connections. If the pruning of these edges results in the isolation of nodes, one can consider these isolated nodes to be less relevant to the network structure than the ones that remain in the backbone. As discussed in Section~\ref{subsec:method-community}, we perform network backboning in two steps. To remove the nodes which are not likely to hold much meaning, we remove the most common words on Twitter from the network if they are also frequently occurring in the corpus, which is equivalent to parsing the original tweets with these words as stop words. We then use the disparity filter or the noise-corrected model to remove low-weight edges. We find that the noise-corrected model removes few edges from the network despite setting the significance level to be very low; in contrast, the disparity filter removes both edges and nodes at a wide range of thresholds. This is unsurprising, as the noise-corrected model gives a higher importance to edges that connect low-degree nodes compared to the disparity filter and is known to be the less restrictive algorithm of the two. In the following discussion, we only consider results obtained using the disparity filter. The disparity filter only begins to take out edges at threshold values $\alpha<0.5$, with a sharp drop in the number of nodes and edges beginning at $\alpha=0.3$. Although similar communities are found for $\alpha=0.3$ up to the most restrictive value $\alpha=0.05$, we find that $\alpha=0.05$ removes all edges with a weight of 1, and consequently produces fewer irrelevant words. We thus present the results corresponding to $\alpha=0.05$ in the rest of the manuscript. More details regarding the differences between different values of $\alpha$ are found in the Supplementary Information.

Depending on the corpus, the edges that are removed at $\alpha=0.05$ are not random (Figure~\ref{fig:hclinton_scorepair_by_threshold}). In the ``favor'' corpus, the vast majority of edges connecting negative words to other negative words have been removed, as well as edges that connect extremely positive words. In contrast, the removal of edges in the ``against'' corpus does not have a clear dependence on the scores. In all corpora, the dominance of neutral-neutral connections still holds, albeit at a tighter range.

\begin{figure*}[ht!]
    \centering
    \includegraphics[width=\textwidth]{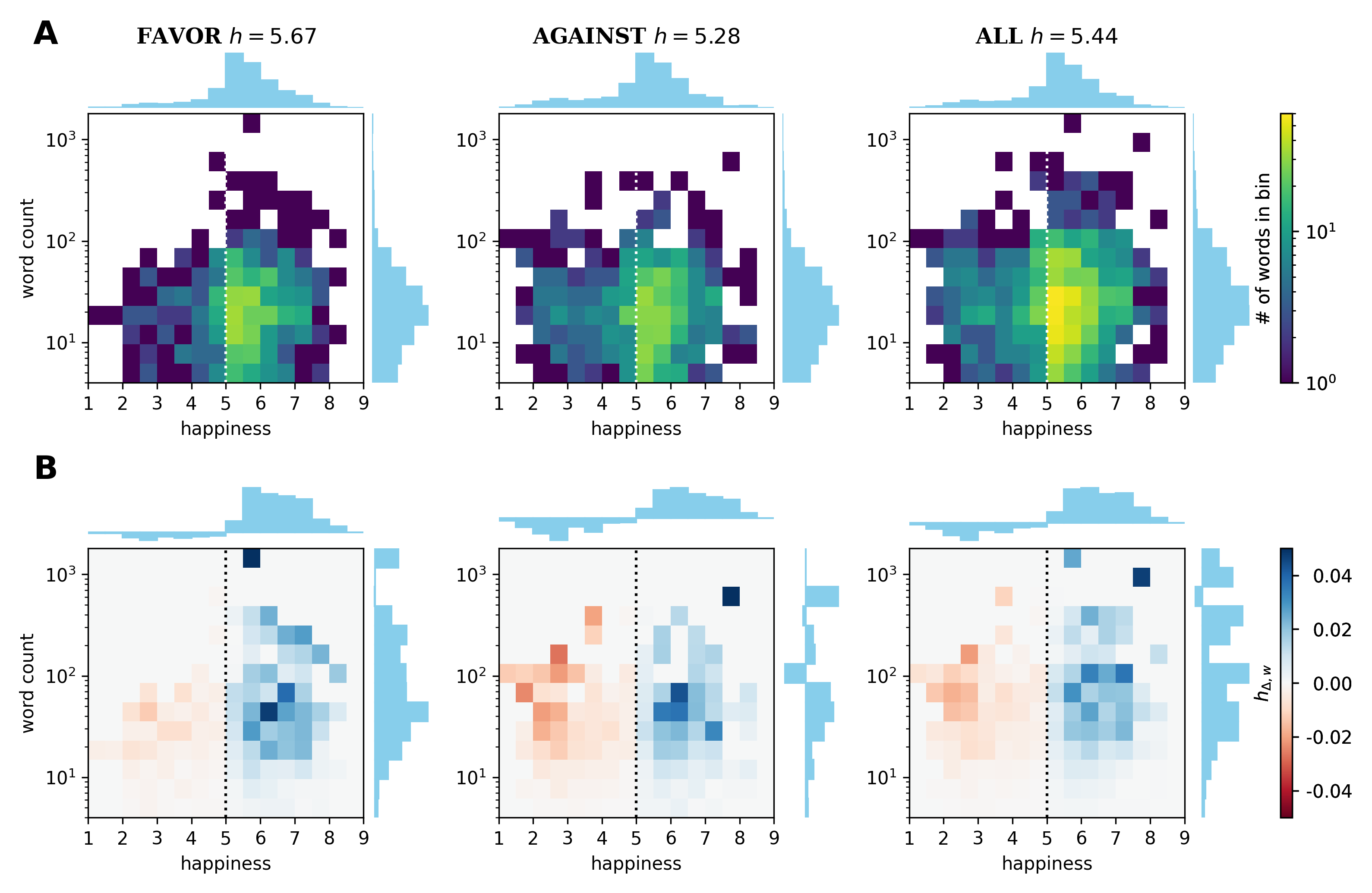}
    \caption{\textbf{Distribution of words and their contributions to the deviation from neutrality in terms of their word counts and happiness scores after network backboning.} A 2D histogram of (a) the word counts and (b) the deviation from neutrality $h_{\Delta,w}$ of the words for each network backbone at $\alpha=0.05$.}
    \label{fig:hclinton_backbone_scorecontrib_dist_v2}
\end{figure*}

Figure~\ref{fig:hclinton_backbone_scorecontrib_dist_v2} shows the histogram of the contribution to the deviation from neutrality of each word, $h_{\Delta,w}$ for $\alpha=0.05$, the most restrictive of the values tested. Compared to the complete network (Figure~\ref{fig:hclinton_scorecontrib_count_dist_v2}), the ``favor'' corpus lost a significant number of words with happiness scores below $h=5$. On the contrary, the ``against'' corpus did not lose as many negative words, although many positive words remained. There is also a clearer difference in the score profiles of the ``favor'' and ``against'' backbone networks, and this was attained not through the use of score cutoffs but by using the co-occurrence network structure itself. Aside from this, there are now fewer words in the backbone that have low word counts, increasing the overall influence of non-neutral words that occur with higher frequency. The ``all'' corpus, being a combination of the two, has similarities to both. To see if we can extract the properties of the ``favor'' and ``against'' stances in the ``all'' corpus, we turn to community detection.

\subsection{Community detection and characterization}
\label{subsec:results-community}

We have seen that by removing commonly used words on Twitter, a number of high-degree neutral words were removed, increasing the proportion of positive and negative words in the corpus. Using the disparity filter further removed low-weight edges. With more restrictive thresholds, edges connecting negative words to other negative words are removed in the ``favor'' corpus, while no distinct patterns are seen in the ``against'' and ``all'' corpora. Whereas negative words were removed from both the ``favor'' and ``against'' corpora, the ``favor'' corpus lost more negative words, resulting in an amplified difference between the average happiness score of the two corpora.

We now investigate whether there is meaningful community structure in the co-occurrence network, and whether this structure can be used to uncover and isolate mixed sentiments in the corpora, particularly in the ``all'' corpus where we expect opposing sentiments to exist. We find that different values of the disparity filter threshold $\alpha$ all give consistent communities that fit a given theme, with more restrictive values of $\alpha$ keeping the communities that can be attributable to known events and reducing the number of communities with words that are less coherent. The following results were obtained using the Louvain algorithm on the backbone derived using $\alpha=0.05$.

\subsubsection{Characterizing communities}
\label{subsubsec:results-community-wordcloud}

\begin{figure*}[ht!]
    \centering
    \includegraphics[width=.82\textwidth]{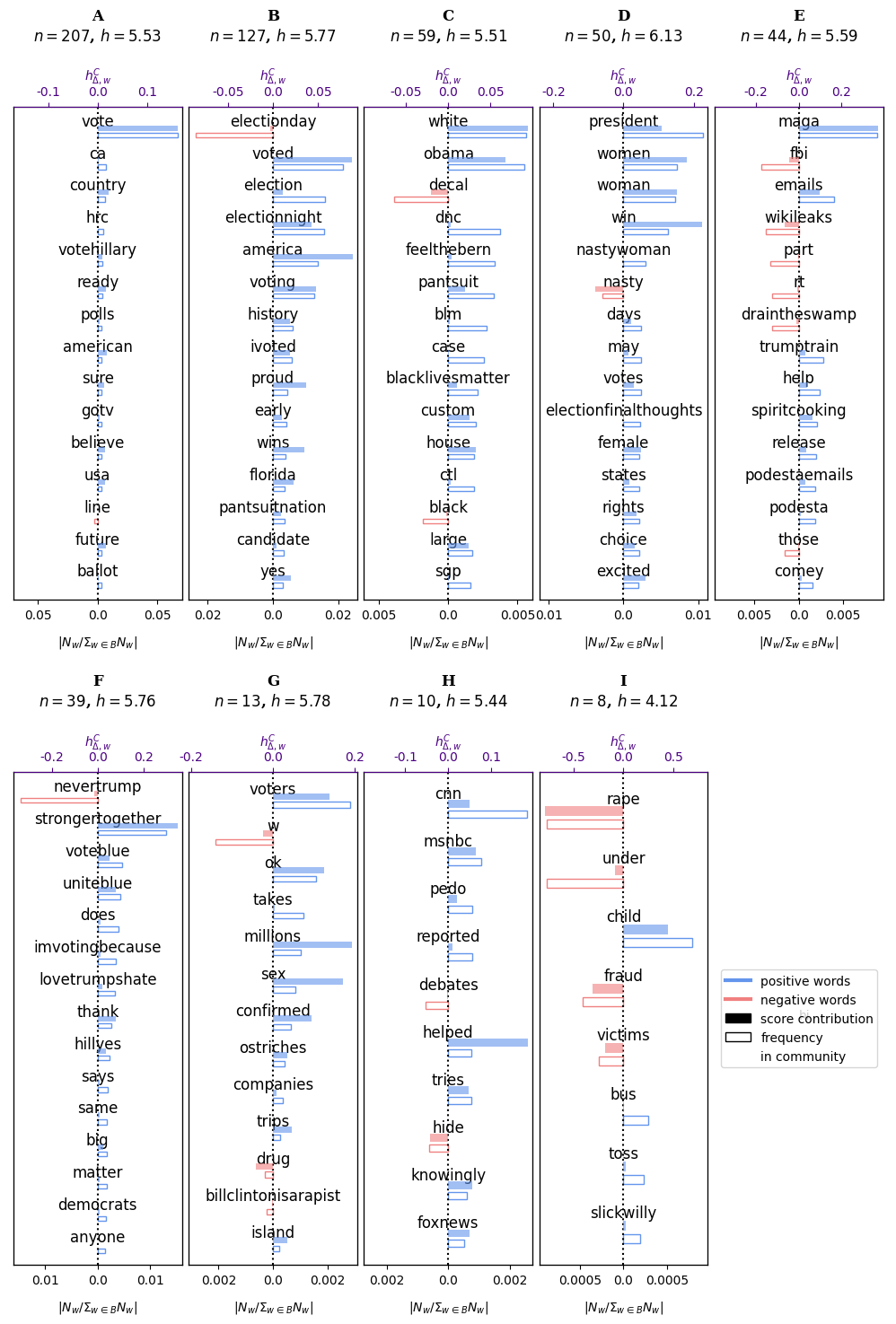}
    \caption{\textbf{Word bars obtained for the top 9 communities with the most number of nodes for the ``FAVOR'' tweets.} The shaded bars give the deviation from normality ($h^{C}_{\Delta,w}$, top x-axis) as computed by Equation~\ref{eq:score_contrib} for the community. The unfilled bars give the relative frequency of the word compared to the sum of all the counts of the words in the backbone $B$ ($N_w/\sum_{w\in B}N_w$, bottom x-axis). Note that the bottom x-axis is symmetric about 0 to always make the unfilled bars appear directly below the shaded bars. The number of nodes and the average scores for each community are given above each plot.}
    \label{fig:hclinton_wordbar_FAVOR}
\end{figure*}

\begin{figure*}[ht!]
    \centering
    \includegraphics[width=.82\textwidth]{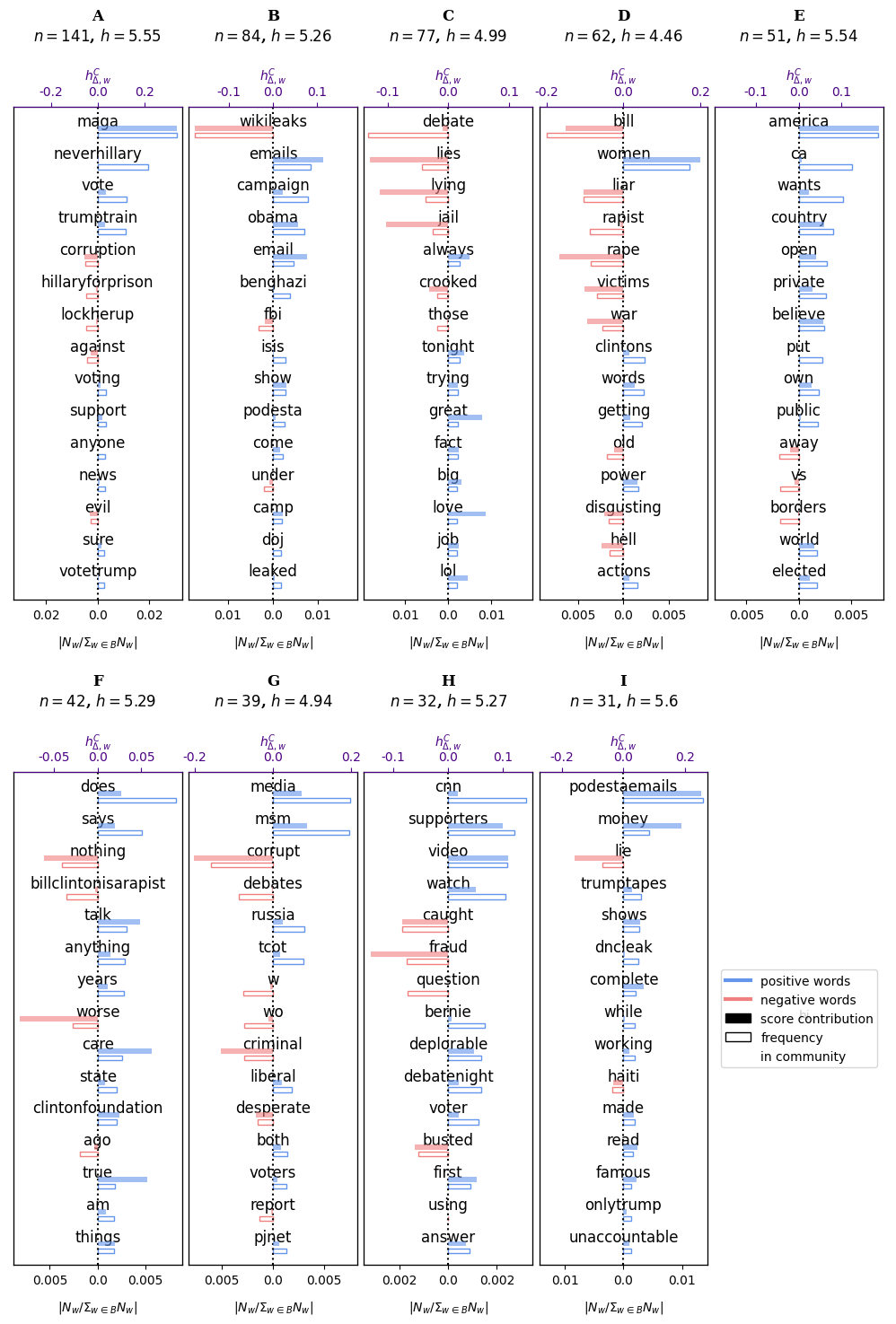}
    \caption{\textbf{Word bars obtained for the top 9 communities with the most number of nodes for the ``AGAINST'' tweets.} The shaded bars give the deviation from normality ($h^{C}_{\Delta,w}$, top x-axis) as computed by Equation~\ref{eq:score_contrib} for the community. The unfilled bars give the relative frequency of the word compared to the sum of all the counts of the words in the backbone $B$ ($N_w/\sum_{w\in B}N_w$, bottom x-axis). Note that the bottom x-axis is symmetric about 0 to always make the unfilled bars appear directly below the shaded bars. The number of nodes and the average scores for each community are given above each plot.}
    \label{fig:hclinton_wordbar_AGAINST}
\end{figure*}

\begin{figure*}[ht!]
    \centering
    \includegraphics[width=.82\textwidth]{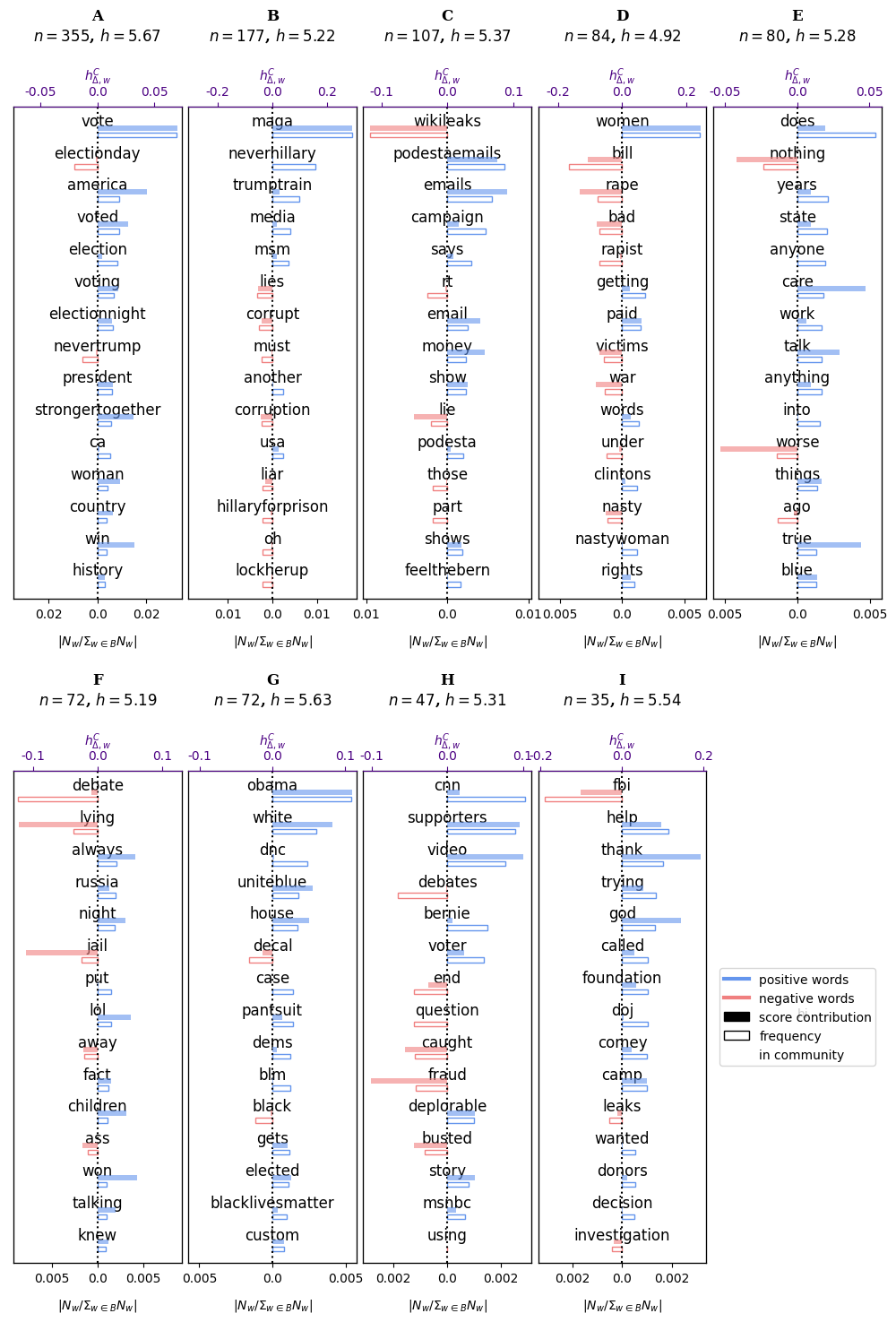}
    \caption{\textbf{Word bars obtained for the top 9 communities with the most number of nodes for the ``ALL'' tweets.} The shaded bars give the deviation from normality ($h^{C}_{\Delta,w}$, top x-axis) as computed by Equation~\ref{eq:score_contrib} for the community. The unfilled bars give the relative frequency of the word compared to the sum of all the counts of the words in the backbone $B$ ($N_w/\sum_{w\in B}N_w$, bottom x-axis). Note that the bottom x-axis is symmetric about 0 to always make the unfilled bars appear directly below the shaded bars. The number of nodes and the average scores for each community are given above each plot.}
    \label{fig:hclinton_wordbar_ALL}
\end{figure*}

Figures~\ref{fig:hclinton_wordbar_FAVOR} to~\ref{fig:hclinton_wordbar_ALL} show the words with the highest word counts in each of the biggest communities (note that for the backbone network, the degree is no longer necessarily correlated with the word counts), their contributions to the deviation from normality of the average score of the community, and their word counts in the corpus.  We see that the biggest communities consist of related words and form a consistent narrative.

In the ``favor'' corpus (Figure~\ref{fig:hclinton_wordbar_FAVOR}), we see that the communities A and B are about the elections; D relates to being a woman president. We note that despite (or rather, due to) ``nasty woman'' being a term used by Donald Trump to denigrate Hillary Clinton, the phrase has been used as a rallying cry by feminists to support Clinton~\cite{gray_how_2016}, so it is not a surprise to find it with words that are likely to favor her. On the other hand, Community E is a community with Donald Trump's ``MAGA'' slogan (short for ``Make America Great Again''). As expected, it also contains other words and hashtags that relate to scandals concerning Hillary Clinton, but the corresponding word counts are less than those in the biggest communities A and B. Community C contains words that are related to the Democratic party, with the word ``white'' likely often used in the context of both the ``White House'' and the Black Lives Matter movement.

In each of the larger communities (except for community E), the most frequently occurring words are the words that we expect to be associated with a favorable stance towards Hillary Clinton.  The smaller communities contain words that do not necessarily conform to a theme, although we note that there are a number of isolated communities that contain infrequently used words. Examples of these are: ``louis'' and ``ck'' (Louis C.K. is a comedian); ``west'' and ``coast'' (West Coast is used to refer to the west coast of the United States); and ``lady'' and ``gaga'' (Lady Gaga is a singer). Overall, the consistency of our results with the expectations from our domain knowledge supports the validity of the backboning and community detection algorithms used.

Similar patterns can be found in the ``against'' corpus (Figure~\ref{fig:hclinton_wordbar_AGAINST}). The biggest community, A, is characterized by words that are for voting Trump rather than Clinton (``maga'', ``vote'', ``trumptrain'') and also include hashtags that are against Hillary such as ``neverhillary'',  ``hillaryforprison'', and ``lockherup''). Community B contains words relating to scandals against Hillary Clinton (``wikileaks'', ``emails'', ``benghazi'', ``collusion'', ``podesta''), while community D contains the words ``bill'', ``women'', and ``rapist'', which are likely related to rape allegations against Hillary Clinton's husband Bill Clinton. Some smaller communities have words that are related to a theme, such as community G (``msm'', ``media'', and ``debates'') and community H (``cnn'', ``video'', and ``caught'' as well as ``anderson'', ``cooper'', ``msnbc'') , but there are also words in these communities that are not related.

In the combined corpus (Figure~\ref{fig:hclinton_wordbar_ALL}), we see communities similar to the dominant ones in the ``favor'' and ``against'' corpora. Community A is about voting, community B contains words that are either against Hillary or for Donald Trump, community C contains words relating to scandals involving Hillary Clinton, and community D contains words that relate to the rape allegations against Bill Clinton. Some themes in smaller communities also exist. In community G, we see words relating to other Democrats and the Black Lives Matter movement, while community H contains words about the media (``cnn'', ``msnbc'', as well as ``nbc'', ``foxnews'',  ``anderson'', ``cooper'', ``bernie'' and ``sanders'', referring to Bernie Sanders who was Hillary Clinton's main opponent in the Democrat primaries).

\subsubsection{Communities and scores}
\label{subsubsec:results-community-scores}

We have shown how disjoint communities in the network correspond to certain themes, with themes that are for or against Hillary Clinton clearly separated in the different corpora. We now explore whether communities of different themes differ in terms of the scores of the words they contain.

\begin{figure*}[ht!]
    \centering
    \includegraphics[width=\textwidth]{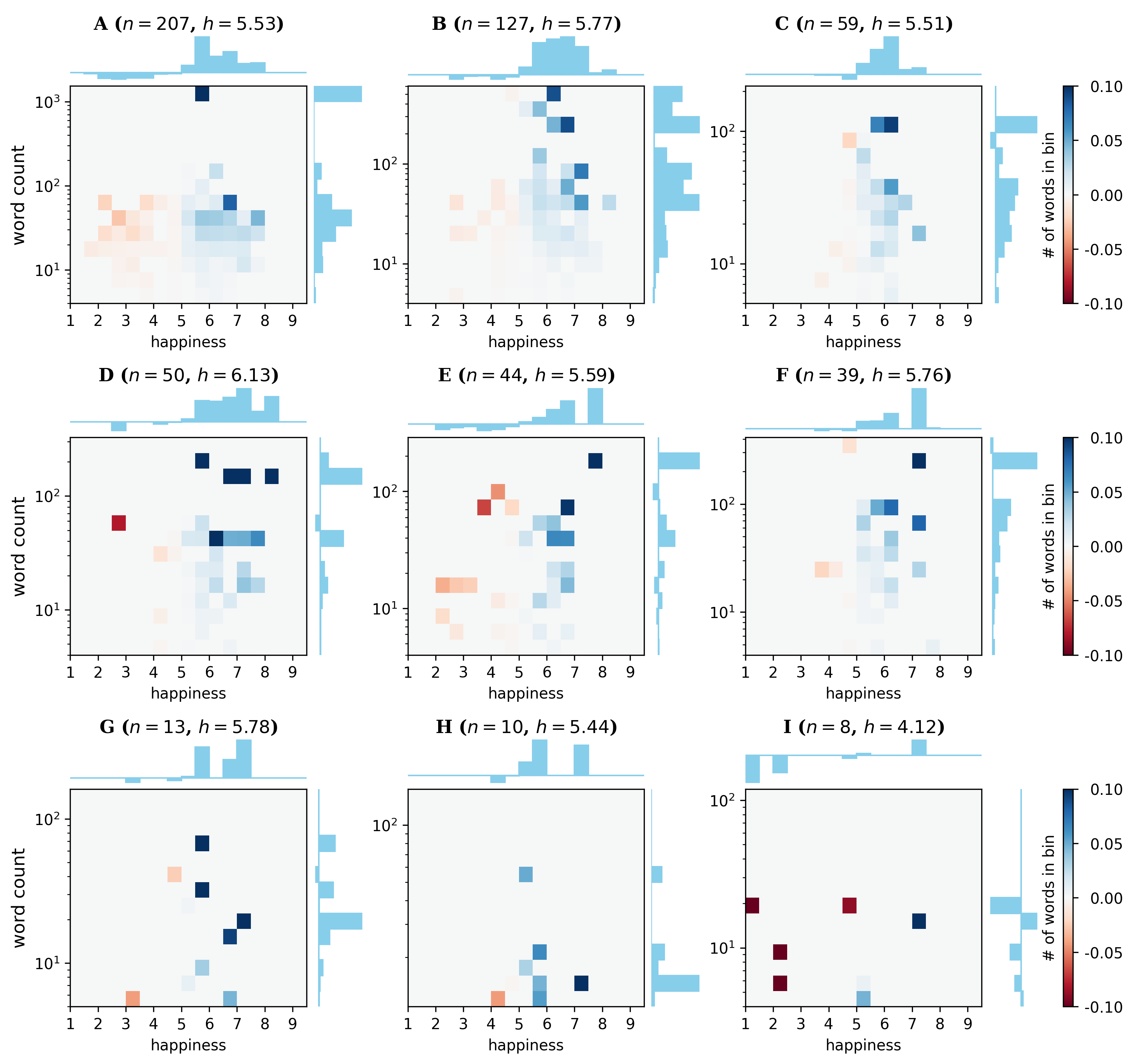}
    \caption{\textbf{Score contributions in communities for the ``FAVOR'' corpus.} A 2D histogram of the score contributions of words as a function of their scores and word counts for the 9 biggest communities in the ``FAVOR'' corpus. The average happiness score and number of nodes for each community is given above each plot.}
    \label{fig:hclinton_comm_scorecontrib_FAVOR}
\end{figure*}

The biggest communities in ``favor'' corpus are dominated by positive contributions from normality (Figure~\ref{fig:hclinton_comm_scorecontrib_FAVOR}, while notable negative words include ``nasty'' (community D), ``wikileaks'' (community E), and ``rape'' (community I). As discussed earlier, the word ``nasty'' (in the context of ``nasty woman''), though inherently negative, has been used by supporters of Hillary Clinton. Closer examination of tweets reveal that although the hashtag ``\#imwithher'' is known mainly to express opinions favorable to Hillary Clinton, it is also hijacked by those who express negative opinions about her, and we find that community detection on the backbone of the co-occurrence network was able to isolate these negative themes.

\begin{figure*}[ht!]
    \centering
    \includegraphics[width=\textwidth]{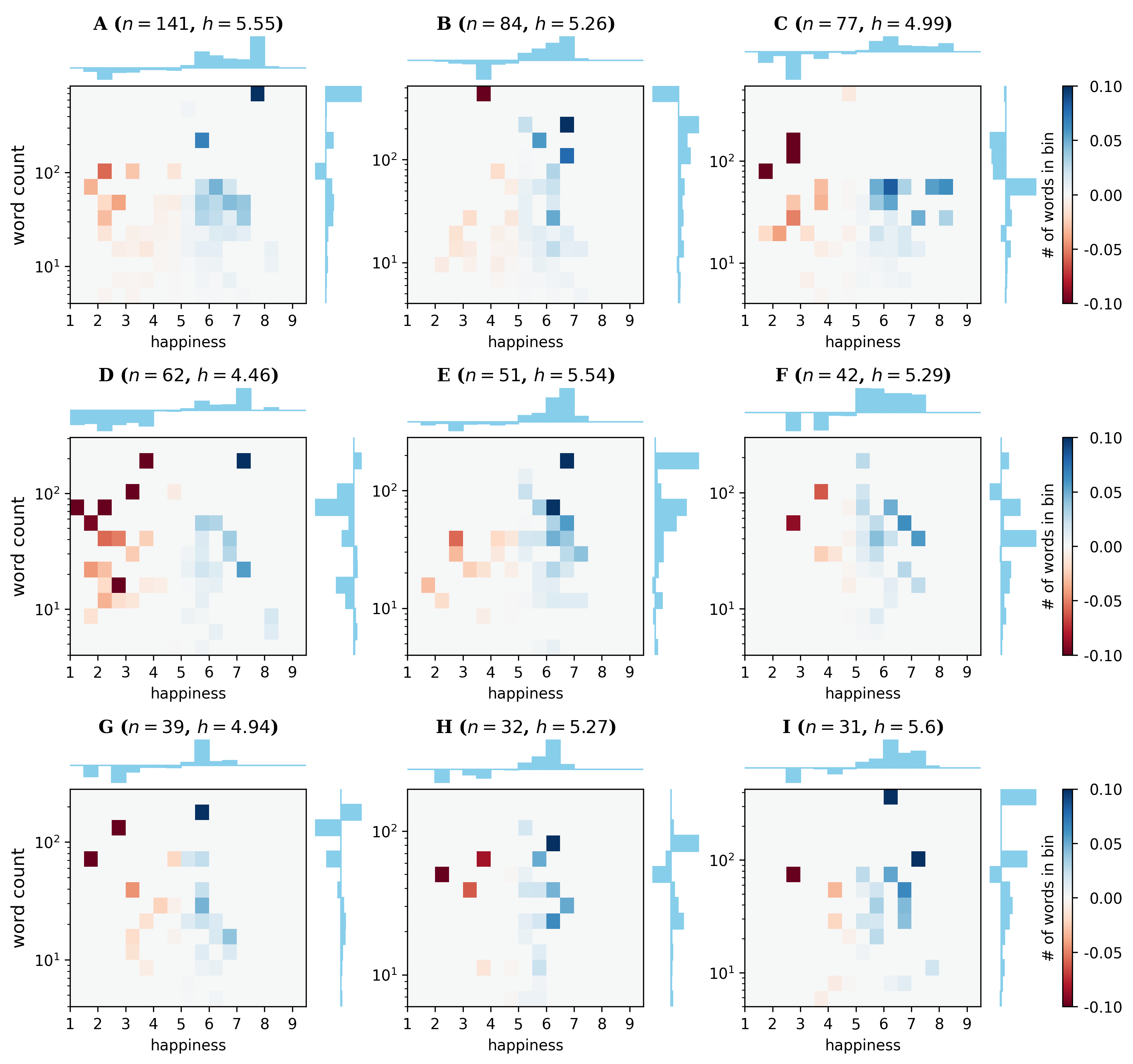}
    \caption{\textbf{Score contributions in communities for the ``AGAINST'' corpus.}A 2D histogram of the score contributions of words as a function of their scores and word counts for the 9 biggest communities in the ``AGAINST'' corpus. The average happiness score for each community is given above each plot.}
    \label{fig:hclinton_comm_scorecontrib_AGAINST}
\end{figure*}

The biggest community (community A) in the ``against'' corpus is characterized by words of support for Donald Trump, primarily by the acronym ``maga'' (``make America great again''), Donald Trump's slogan in the 2016 elections (Figure~\ref{fig:hclinton_comm_scorecontrib_AGAINST}. Community B contains words relating to the scandals involving Hillary Clinton (``wikileaks''). It also contains ``emails'' which has a positive score in the labMT dataset, despite it having a negative connotation in the context of the 2016 US presidential elections. Community C has the negative words ``lies'', ``lying'', and ``jail'', but also the positive words ``love'' and ``great''. We note that although ``love'' and ``great'' are some of the most commonly used words on Twitter, they are not as frequently used in this corpus and were therefore not filtered out. Closer examination of the raw tweets show that these positive words are used in a negative or sarcastic sense against Hillary Clinton. Community D has an extremely low negative score due to the words associated with ``rape''. The word ``bill'', which was mostly used in the context of ``Bill Clinton'', has a score of 3.64, as it was probably interpreted by the labMT respondents to mean ``invoice'', as in ``medical bill'' or ``electric bill''. Regardless, even if ``bill'' were not scored, we see that the words in community D were mainly negative, consistent with its association with ``rape''. A notable exception is the word ``women'', which has a happiness score of 7.12, even though it is used often with the word ``rape'' in this context.

\begin{figure*}[ht!]
    \centering
    \includegraphics[width=\textwidth]{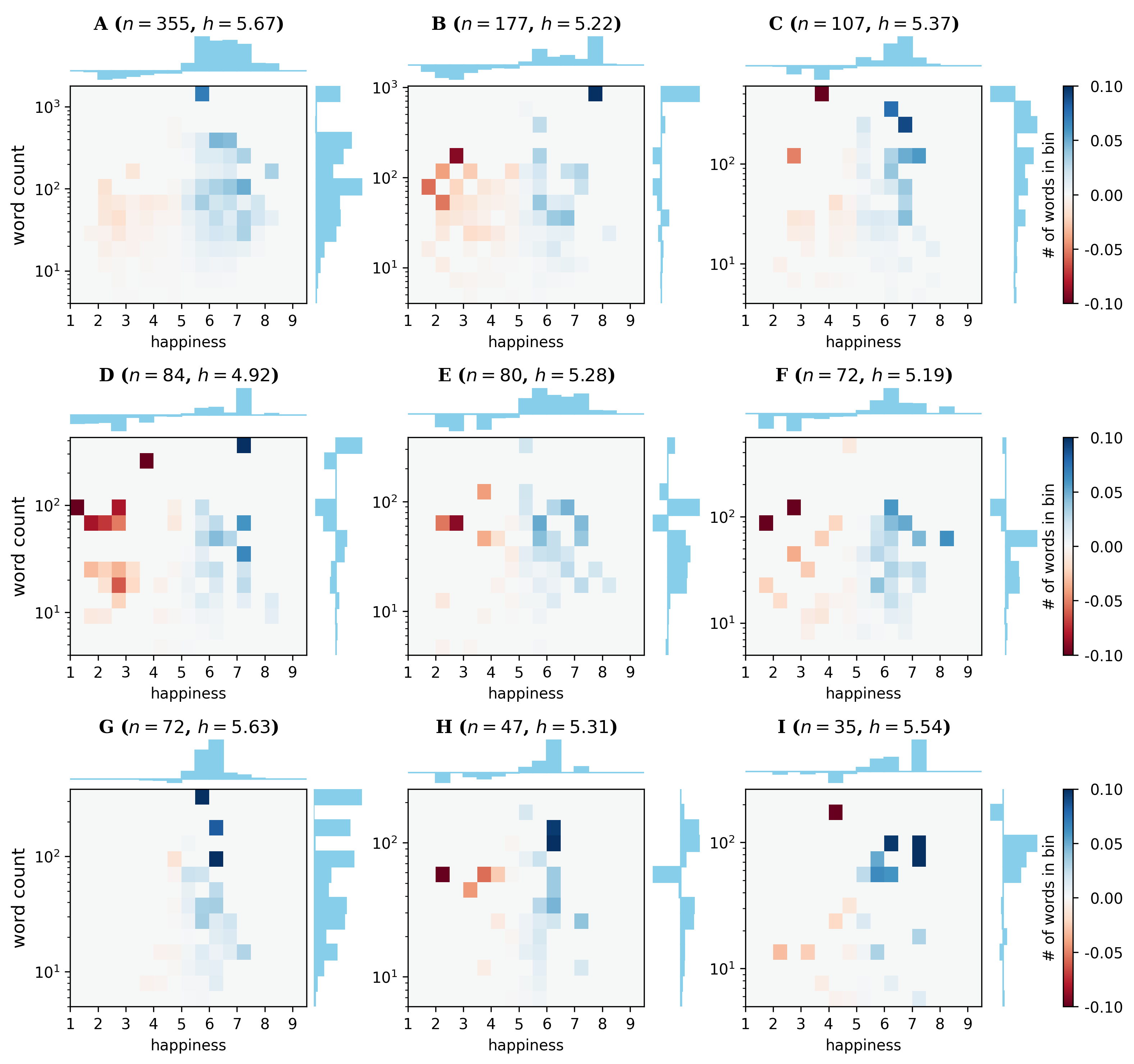}
    \caption{\textbf{Score contributions in communities for the ``ALL'' corpus.} A 2D histogram of the score contributions of words as a function of their scores and word counts for the 9 biggest communities in the ``ALL'' corpus. The average happiness score for each community is given above each plot.}
    \label{fig:hclinton_comm_scorecontrib_ALL}
\end{figure*}

The biggest community (community A) in the combination of the two corpora (``all'') is mainly positive (Figure~\ref{fig:hclinton_comm_scorecontrib_ALL}, which is consistent with the fact that the most frequently used words in this community are in support of Hillary Clinton. Community B has a fair share of negative words that are used against Hillary Clinton (``lies'', ``corrupt'', ``criminal'') but are also used with Trump's slogan ``maga'', which carries a positive score. Community C, being about particular scandals such as ``wikileaks'', has fewer negative words. Further, some words that are used in a negative sense in this context are in general either positive or neutral words (``emails'', ``spiritcooking''). Community D contains many negative words associated with ``rape'', but also positive words such as ``women'', which in turn commonly occurs with ``nasty'' (in the context of ``nasty women''). Communities E and F both have dominant negative words, but the words in these communities do not share coherent themes.

\begin{figure*}[ht!]
    \centering
    \includegraphics[width=\textwidth]{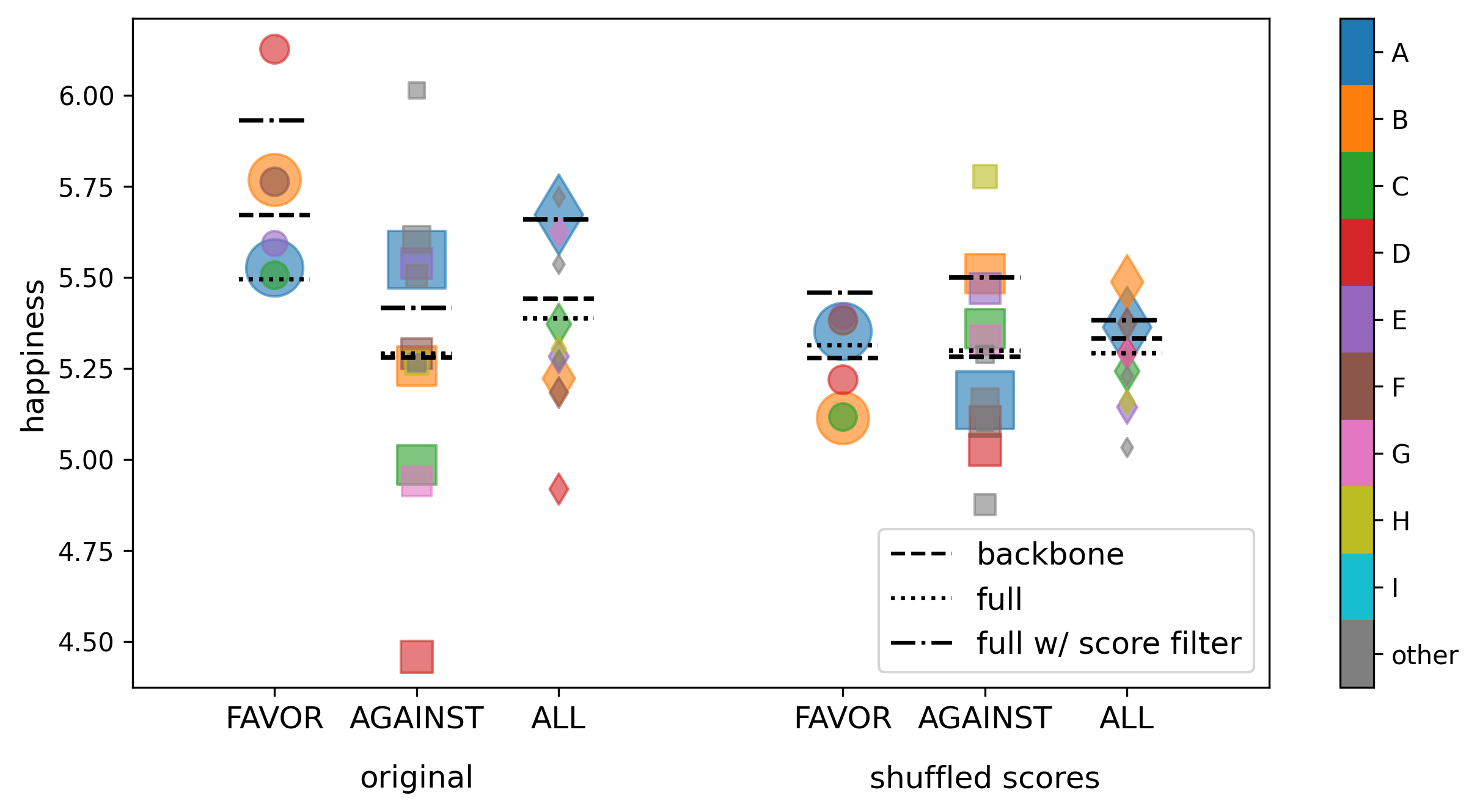}
    \caption{\textbf{Average happiness scores for each community.} The average happiness scores weighted by the word count for each community for communities with at least 15 words are shown in the plot. The set of communities on the left is for the original backboned network, while the one on the right is for the same network but with the scores shuffled among the nodes. Whereas the ``favor'', ``against'', and ``all'' communities in the original backboned network show different ranges, the shuffled score version gives similar ranges for the communities in the three corpora, indicating that the score profiles found in communities are not random. The dashed line is the average happiness of all the words in the backbone, the dotted line is the average happiness of all the words in the raw network, and the dash-dot line gives the average happiness of the raw network if nodes with scores in the range (4,6) are excluded. The size of each symbol is proportional to the total word counts of all the words in that community; note that with the backboning this is no longer necessarily correlated with the degree. The colors indicate the community labels; smaller communities than A-I are all labeled ``other''.}
    \label{fig:hclinton_comm_scores}
\end{figure*}

Figure~\ref{fig:hclinton_comm_scores} shows the average happiness scores weighted by the word count for each community with at least 15 nodes. For comparison, we also include the average scores per community obtained if the scores were shuffled across the unfiltered network, but the network structure (and thus the communities) were retained. In the shuffled version, the range of the community scores are similar across stances, which is not the case in the original network, indicating that the score profiles found for the communities are not random. We also include in the figure the average happiness scores computed if no network backboning was implemented, but if words close to neutral ($4<h<6$) were disregarded, as is customary in studies using happiness scores~\cite{dodds_temporal_2011,dodds_human_2015}. While removing these words increases the scores across all corpora, reflecting the dominance of positive words in text~\cite{dodds_human_2015}, network backboning only increased the scores in the ``favor'' corpus and not in the ``against'' corpus, resulting in a moderate increase in the ``all'' corpus compared to the average happiness score derived from the complete network.

\FloatBarrier

\section{Discussion and summary}
\label{sec:conclusion}

From the word co-occurrence network, we examined how the co-occurrence structure of words in political tweets is related to the word happiness scores. In particular, we looked at three corpora that relate to the American politician Hillary Clinton: one collected using the hashtag ``\#imwithher'', which is associated with being in favor of Clinton; another collected using the hashtag ``\#crookedhillary'', which is associated with being against her; and a combination of both corpora. The degree, node strength, word count and tweet count distributions are all heavy-tailed, consistent with what is expected from Zipf's law ~\cite{ryland_williams_zipfs_2015,williams_text_2015}. Although it is expected that neutral words would dominate the corpus without any backboning or filtering, we find little assortative mixing outside the neutral range. Contrary to our expectation, while positive or negative words are more likely to co-occur with neutral words, positive or negative words are not more likely to co-occur with words of similar polarity. A clear demonstration of this is the use of the hashtag ``\#maga'', which on its own exhibits a highly positive sentiment. However, as a slogan for Donald Trump, it is also used together with words with a negative sentiment against Hillary Clinton. Further, we compared the score profiles with null models that change the network structure or the scores, and find that network structure plays a much smaller role in the score profile than the score distribution.

Given that several words in the corpus do not occur frequently, we extracted a more robust network structure using network backboning. We implemented it in two passes, first by removing frequently occurring words on Twitter, and second, by using the disparity filter to remove edges between words that co-occur less frequently. Performing community detection on the resulting backbone results in well-defined communities that correspond to themes in favor or against the target (Hillary Clinton), even for the ``favor'' or ``against'' corpora that lean more towards only one stance. Effectively, the Louvain algorithm, a non-overlapping community detection model, functions like a topic model when applied to the word co-occurrence network when co-occurrence is taken at a tweet level.

The average happiness scores of the communities also correspond to the theme that they are associated with. Communities dominated by words in favor of Hillary Clinton tend to be have higher happiness scores, while those dominated by words against her tend to have lower happiness scores. A few context-specific words whose scores do not reflect their meaning are ``nastywoman'', ``nasty'', and the name ``bill'', but even if these words are disregarded, we still see similar behavior for the rest of the words in their communities. We also note that both positive and negative words are present in each community. Notably, even in the most negative communities in the ``against'' corpus, we find positive words, reaffirming the positivity bias of human language~\cite{dodds_human_2015} found even in negative statements~\cite{aithal_positivity_2021}.

In summary, we found that raw co-occurrence networks in tweets are dominated by neutral words as well as connections between them, and no score homophily is observed. Network backboning reduces the influence of neutral words, but does not uncover assortative mixing. However, when split into communities, the average happiness score of the community corresponds with the theme. Further, opposing themes are found, not just when the opposing tweets are equal in number (as in the ``all'' corpus), but also when an opposing theme is in the minority. This indicates that the word co-occurrence network with backboning and community detection can be used to detect the presence of opposing sentiments within a corpus.

We have only tested our methods on political tweets, and also for a limited subset. Because generating the word co-occurrence network does not involve any filters, applying our analysis to a larger dataset, such as an entire day's worth of tweets or a more popular anchor such as all tweets mentioning the words ``Trump" or ``BTS", which have extremely high word counts on Twitter~\cite{dodds_fame_2021}, would be computationally expensive. Implementing a suitable filter in the network generation step may be necessary and is an avenue for future work. We would also like to expand our work in the future to analyze and compare tweets from different domains, as well as include higher order n-grams in our analysis.

\FloatBarrier


\acknowledgments
The authors are grateful for the computing resources provided by the Vermont Advanced Computing Core 
and financial support from the Massachusetts Mutual Life Insurance Company.
Computations were performed on the Vermont Advanced Computing Core supported in part by NSF award No. OAC-1827314. M.I.F. is also grateful to Takayuki Hiraoka for helpful discussions.





\clearpage

\newwrite\tempfile
\immediate\openout\tempfile=startsupp.txt
\immediate\write\tempfile{\thepage}
\immediate\closeout\tempfile

\renewcommand{\thefigure}{S\arabic{figure}}
\renewcommand{\thetable}{S\arabic{table}}
\setcounter{figure}{0}
\setcounter{table}{0}
\newpage
\onecolumngrid
\section*{Supplementary Material}
\subsection{Removing hashtag hijacking}

When we performed community detection on all the tweets, we found a community with words unrelated to Hillary Clinton or the elections, such as ``santa'' and ``christmas''. Upon checking the raw tweets with these words, we found that they all contained the hashtag \texttt{\#smptweettest}. All tweets with this hashtag were removed in the dataset analyzed in the manuscript.

\subsection{Most frequently used words removed in network backboning}
Table~\ref{tab:removed_words} shows the words removed in the initial backboning step by comparing the words that consistently made it to the top 400 1-grams in Twitter for 100 random days and the top 200 case-insensitive 1-grams in each of the ``favor'', ``against'', and ``all'' corpora.
\begin{longtable*}{|l|c|c|c|l|l|c|c|c|}
\caption{Words from the intersection of the top 400 1-grams in Twitter for 100 random days, and whether or not they were also present in the most frequently occurring 200 case-insensitive words in each of the ``favor'', ``against'', and ``all'' corpora. Note that the top 400 1-grams from Twitter is case-sensitive and includes non-alphabet symbols, while the words in this table are case-insensitive and only contain characters from the English alphabet. Words with a check mark were removed in the initial backboning step.}
\label{tab:removed_words} \\
\cline{1-4} \cline{6-9}
\textbf{word}     & FAVOR & AGAINST & ALL   &  & \textbf{word}      & FAVOR & AGAINST & ALL   \\ \cline{1-4} \cline{6-9} 
\textbf{a}        & \checkmark  & \checkmark    & \checkmark  &  & \textbf{many}      &  & \checkmark    & \checkmark  \\ \cline{1-4} \cline{6-9} 
\textbf{about}    & \checkmark  & \checkmark    & \checkmark  &  & \textbf{me}        & \checkmark  & \checkmark    & \checkmark  \\ \cline{1-4} \cline{6-9} 
\textbf{after}    & \checkmark  & \checkmark    & \checkmark  &  & \textbf{miss}      &  &    &  \\ \cline{1-4} \cline{6-9} 
\textbf{again}    & \checkmark  & \checkmark    & \checkmark  &  & \textbf{more}      & \checkmark  & \checkmark    & \checkmark  \\ \cline{1-4} \cline{6-9} 
\textbf{all}      & \checkmark  & \checkmark    & \checkmark  &  & \textbf{morning}   &  &    &  \\ \cline{1-4} \cline{6-9} 
\textbf{always}   &  &    &  &  & \textbf{most}      & \checkmark  & \checkmark    & \checkmark  \\ \cline{1-4} \cline{6-9} 
\textbf{am}       & \checkmark  &    & \checkmark  &  & \textbf{much}      & \checkmark  & \checkmark    & \checkmark  \\ \cline{1-4} \cline{6-9} 
\textbf{an}       & \checkmark  & \checkmark    & \checkmark  &  & \textbf{my}        & \checkmark  & \checkmark    & \checkmark  \\ \cline{1-4} \cline{6-9} 
\textbf{and}      & \checkmark  & \checkmark    & \checkmark  &  & \textbf{need}      & \checkmark  & \checkmark    & \checkmark  \\ \cline{1-4} \cline{6-9} 
\textbf{any}      &  & \checkmark    & \checkmark  &  & \textbf{never}     & \checkmark  & \checkmark    & \checkmark  \\ \cline{1-4} \cline{6-9} 
\textbf{are}      & \checkmark  & \checkmark    & \checkmark  &  & \textbf{new}       & \checkmark  & \checkmark    & \checkmark  \\ \cline{1-4} \cline{6-9} 
\textbf{as}       & \checkmark  & \checkmark    & \checkmark  &  & \textbf{next}      & \checkmark  &    &  \\ \cline{1-4} \cline{6-9} 
\textbf{at}       & \checkmark  & \checkmark    & \checkmark  &  & \textbf{night}     & \checkmark  &    &  \\ \cline{1-4} \cline{6-9} 
\textbf{back}     & \checkmark  & \checkmark    & \checkmark  &  & \textbf{no}        & \checkmark  & \checkmark    & \checkmark  \\ \cline{1-4} \cline{6-9} 
\textbf{bad}      &  & \checkmark    &  &  & \textbf{not}       & \checkmark  & \checkmark    & \checkmark  \\ \cline{1-4} \cline{6-9} 
\textbf{be}       & \checkmark  & \checkmark    & \checkmark  &  & \textbf{now}       & \checkmark  & \checkmark    & \checkmark  \\ \cline{1-4} \cline{6-9} 
\textbf{because}  & \checkmark  & \checkmark    & \checkmark  &  & \textbf{of}        & \checkmark  & \checkmark    & \checkmark  \\ \cline{1-4} \cline{6-9} 
\textbf{been}     & \checkmark  & \checkmark    & \checkmark  &  & \textbf{off}       & \checkmark  & \checkmark    & \checkmark  \\ \cline{1-4} \cline{6-9} 
\textbf{before}   & \checkmark  &    &  &  & \textbf{on}        & \checkmark  & \checkmark    & \checkmark  \\ \cline{1-4} \cline{6-9} 
\textbf{being}    & \checkmark  & \checkmark    & \checkmark  &  & \textbf{one}       & \checkmark  & \checkmark    & \checkmark  \\ \cline{1-4} \cline{6-9} 
\textbf{best}     & \checkmark  &    &  &  & \textbf{only}      & \checkmark  & \checkmark    & \checkmark  \\ \cline{1-4} \cline{6-9} 
\textbf{better}   & \checkmark  &    &  &  & \textbf{or}        & \checkmark  & \checkmark    & \checkmark  \\ \cline{1-4} \cline{6-9} 
\textbf{but}      & \checkmark  & \checkmark    & \checkmark  &  & \textbf{other}     &  &    &  \\ \cline{1-4} \cline{6-9} 
\textbf{by}       & \checkmark  & \checkmark    & \checkmark  &  & \textbf{our}       & \checkmark  & \checkmark    & \checkmark  \\ \cline{1-4} \cline{6-9} 
\textbf{can}      & \checkmark  & \checkmark    & \checkmark  &  & \textbf{out}       & \checkmark  & \checkmark    & \checkmark  \\ \cline{1-4} \cline{6-9} 
\textbf{cant}     &  &    &  &  & \textbf{over}      & \checkmark  & \checkmark    & \checkmark  \\ \cline{1-4} \cline{6-9} 
\textbf{come}     & \checkmark  &    & \checkmark  &  & \textbf{people}    & \checkmark  & \checkmark    & \checkmark  \\ \cline{1-4} \cline{6-9} 
\textbf{could}    & \checkmark  & \checkmark    & \checkmark  &  & \textbf{please}    & \checkmark  &    & \checkmark  \\ \cline{1-4} \cline{6-9} 
\textbf{day}      & \checkmark  &    & \checkmark  &  & \textbf{real}      &  & \checkmark    & \checkmark  \\ \cline{1-4} \cline{6-9} 
\textbf{days}     &  &    &  &  & \textbf{really}    & \checkmark  & \checkmark    & \checkmark  \\ \cline{1-4} \cline{6-9} 
\textbf{did}      & \checkmark  & \checkmark    & \checkmark  &  & \textbf{right}     & \checkmark  & \checkmark    & \checkmark  \\ \cline{1-4} \cline{6-9} 
\textbf{do}       & \checkmark  & \checkmark    & \checkmark  &  & \textbf{rt}        &  & \checkmark    &  \\ \cline{1-4} \cline{6-9} 
\textbf{doing}    &  &    &  &  & \textbf{said}      & \checkmark  & \checkmark    & \checkmark  \\ \cline{1-4} \cline{6-9} 
\textbf{done}     & \checkmark  & \checkmark    & \checkmark  &  & \textbf{same}      &  & \checkmark    &  \\ \cline{1-4} \cline{6-9} 
\textbf{dont}     &  &    &  &  & \textbf{say}       & \checkmark  & \checkmark    & \checkmark  \\ \cline{1-4} \cline{6-9} 
\textbf{down}     & \checkmark  & \checkmark    & \checkmark  &  & \textbf{see}       & \checkmark  & \checkmark    & \checkmark  \\ \cline{1-4} \cline{6-9} 
\textbf{even}     & \checkmark  & \checkmark    & \checkmark  &  & \textbf{she}       & \checkmark  & \checkmark    & \checkmark  \\ \cline{1-4} \cline{6-9} 
\textbf{ever}     & \checkmark  & \checkmark    & \checkmark  &  & \textbf{shit}      &  &    &  \\ \cline{1-4} \cline{6-9} 
\textbf{every}    &  & \checkmark    & \checkmark  &  & \textbf{should}    & \checkmark  & \checkmark    & \checkmark  \\ \cline{1-4} \cline{6-9} 
\textbf{everyone} & \checkmark  &    &  &  & \textbf{so}        & \checkmark  & \checkmark    & \checkmark  \\ \cline{1-4} \cline{6-9} 
\textbf{feel}     & \checkmark  &    &  &  & \textbf{some}      & \checkmark  & \checkmark    & \checkmark  \\ \cline{1-4} \cline{6-9} 
\textbf{find}     &  &    &  &  & \textbf{someone}   &  &    &  \\ \cline{1-4} \cline{6-9} 
\textbf{first}    & \checkmark  &    & \checkmark  &  & \textbf{something} &  &    &  \\ \cline{1-4} \cline{6-9} 
\textbf{follow}   &  &    &  &  & \textbf{still}     & \checkmark  & \checkmark    & \checkmark  \\ \cline{1-4} \cline{6-9} 
\textbf{for}      & \checkmark  & \checkmark    & \checkmark  &  & \textbf{stop}      &  & \checkmark    & \checkmark  \\ \cline{1-4} \cline{6-9} 
\textbf{friends}  &  &    &  &  & \textbf{take}      & \checkmark  & \checkmark    & \checkmark  \\ \cline{1-4} \cline{6-9} 
\textbf{from}     & \checkmark  & \checkmark    & \checkmark  &  & \textbf{tell}      &  &    &  \\ \cline{1-4} \cline{6-9} 
\textbf{fuck}     &  &    &  &  & \textbf{than}      & \checkmark  & \checkmark    & \checkmark  \\ \cline{1-4} \cline{6-9} 
\textbf{get}      & \checkmark  & \checkmark    & \checkmark  &  & \textbf{thanks}    &  &    &  \\ \cline{1-4} \cline{6-9} 
\textbf{getting}  &  &    &  &  & \textbf{that}      & \checkmark  & \checkmark    & \checkmark  \\ \cline{1-4} \cline{6-9} 
\textbf{girl}     &  &    &  &  & \textbf{thats}     & \checkmark  &    &  \\ \cline{1-4} \cline{6-9} 
\textbf{give}     &  &    &  &  & \textbf{the}       & \checkmark  & \checkmark    & \checkmark  \\ \cline{1-4} \cline{6-9} 
\textbf{go}       & \checkmark  & \checkmark    & \checkmark  &  & \textbf{their}     & \checkmark  & \checkmark    & \checkmark  \\ \cline{1-4} \cline{6-9} 
\textbf{going}    & \checkmark  & \checkmark    & \checkmark  &  & \textbf{them}      & \checkmark  & \checkmark    & \checkmark  \\ \cline{1-4} \cline{6-9} 
\textbf{gonna}    & \checkmark  &    &  &  & \textbf{then}      & \checkmark  & \checkmark    & \checkmark  \\ \cline{1-4} \cline{6-9} 
\textbf{good}     & \checkmark  & \checkmark    & \checkmark  &  & \textbf{there}     & \checkmark  & \checkmark    & \checkmark  \\ \cline{1-4} \cline{6-9} 
\textbf{got}      & \checkmark  & \checkmark    & \checkmark  &  & \textbf{these}     & \checkmark  & \checkmark    & \checkmark  \\ \cline{1-4} \cline{6-9} 
\textbf{great}    & \checkmark  &    & \checkmark  &  & \textbf{they}      & \checkmark  & \checkmark    & \checkmark  \\ \cline{1-4} \cline{6-9} 
\textbf{guys}     &  &    &  &  & \textbf{thing}     &  & \checkmark    & \checkmark  \\ \cline{1-4} \cline{6-9} 
\textbf{had}      & \checkmark  & \checkmark    & \checkmark  &  & \textbf{things}    &  &    &  \\ \cline{1-4} \cline{6-9} 
\textbf{happy}    &  &    &  &  & \textbf{think}     & \checkmark  & \checkmark    & \checkmark  \\ \cline{1-4} \cline{6-9} 
\textbf{hard}     &  &    &  &  & \textbf{this}      & \checkmark  & \checkmark    & \checkmark  \\ \cline{1-4} \cline{6-9} 
\textbf{has}      & \checkmark  & \checkmark    & \checkmark  &  & \textbf{time}      & \checkmark  & \checkmark    & \checkmark  \\ \cline{1-4} \cline{6-9} 
\textbf{hate}     & \checkmark  &    &  &  & \textbf{to}        & \checkmark  & \checkmark    & \checkmark  \\ \cline{1-4} \cline{6-9} 
\textbf{have}     & \checkmark  & \checkmark    & \checkmark  &  & \textbf{today}     & \checkmark  &    & \checkmark  \\ \cline{1-4} \cline{6-9} 
\textbf{he}       & \checkmark  & \checkmark    & \checkmark  &  & \textbf{tomorrow}  & \checkmark  &    & \checkmark  \\ \cline{1-4} \cline{6-9} 
\textbf{her}      & \checkmark  & \checkmark    & \checkmark  &  & \textbf{tonight}   & \checkmark  &    & \checkmark  \\ \cline{1-4} \cline{6-9} 
\textbf{here}     & \checkmark  & \checkmark    & \checkmark  &  & \textbf{too}       & \checkmark  & \checkmark    & \checkmark  \\ \cline{1-4} \cline{6-9} 
\textbf{him}      & \checkmark  & \checkmark    & \checkmark  &  & \textbf{tweet}     &  &    &  \\ \cline{1-4} \cline{6-9} 
\textbf{his}      & \checkmark  & \checkmark    & \checkmark  &  & \textbf{u}         & \checkmark  & \checkmark    & \checkmark  \\ \cline{1-4} \cline{6-9} 
\textbf{home}     &  &    &  &  & \textbf{up}        & \checkmark  & \checkmark    & \checkmark  \\ \cline{1-4} \cline{6-9} 
\textbf{hope}     & \checkmark  &    & \checkmark  &  & \textbf{us}        & \checkmark  & \checkmark    & \checkmark  \\ \cline{1-4} \cline{6-9} 
\textbf{how}      & \checkmark  & \checkmark    & \checkmark  &  & \textbf{very}      &  &    &  \\ \cline{1-4} \cline{6-9} 
\textbf{i}        & \checkmark  & \checkmark    & \checkmark  &  & \textbf{via}       & \checkmark  & \checkmark    & \checkmark  \\ \cline{1-4} \cline{6-9} 
\textbf{if}       & \checkmark  & \checkmark    & \checkmark  &  & \textbf{wait}      &  &    &  \\ \cline{1-4} \cline{6-9} 
\textbf{ill}      &  &    &  &  & \textbf{wanna}     &  &    &  \\ \cline{1-4} \cline{6-9} 
\textbf{im}       &  &    &  &  & \textbf{want}      & \checkmark  & \checkmark    & \checkmark  \\ \cline{1-4} \cline{6-9} 
\textbf{in}       & \checkmark  & \checkmark    & \checkmark  &  & \textbf{was}       & \checkmark  & \checkmark    & \checkmark  \\ \cline{1-4} \cline{6-9} 
\textbf{into}     &  & \checkmark    &  &  & \textbf{watch}     & \checkmark  &    & \checkmark  \\ \cline{1-4} \cline{6-9} 
\textbf{is}       & \checkmark  & \checkmark    & \checkmark  &  & \textbf{way}       & \checkmark  & \checkmark    & \checkmark  \\ \cline{1-4} \cline{6-9} 
\textbf{it}       & \checkmark  & \checkmark    & \checkmark  &  & \textbf{we}        & \checkmark  & \checkmark    & \checkmark  \\ \cline{1-4} \cline{6-9} 
\textbf{its}      &  &    &  &  & \textbf{well}      & \checkmark  & \checkmark    & \checkmark  \\ \cline{1-4} \cline{6-9} 
\textbf{ive}      &  &    &  &  & \textbf{were}      & \checkmark  & \checkmark    & \checkmark  \\ \cline{1-4} \cline{6-9} 
\textbf{just}     & \checkmark  & \checkmark    & \checkmark  &  & \textbf{what}      & \checkmark  & \checkmark    & \checkmark  \\ \cline{1-4} \cline{6-9} 
\textbf{keep}     & \checkmark  & \checkmark    & \checkmark  &  & \textbf{when}      & \checkmark  & \checkmark    & \checkmark  \\ \cline{1-4} \cline{6-9} 
\textbf{know}     & \checkmark  & \checkmark    & \checkmark  &  & \textbf{where}     &  & \checkmark    & \checkmark  \\ \cline{1-4} \cline{6-9} 
\textbf{last}     & \checkmark  &    & \checkmark  &  & \textbf{who}       & \checkmark  & \checkmark    & \checkmark  \\ \cline{1-4} \cline{6-9} 
\textbf{let}      & \checkmark  & \checkmark    & \checkmark  &  & \textbf{why}       & \checkmark  & \checkmark    & \checkmark  \\ \cline{1-4} \cline{6-9} 
\textbf{life}     &  &    &  &  & \textbf{will}      & \checkmark  & \checkmark    & \checkmark  \\ \cline{1-4} \cline{6-9} 
\textbf{like}     & \checkmark  & \checkmark    & \checkmark  &  & \textbf{with}      & \checkmark  & \checkmark    & \checkmark  \\ \cline{1-4} \cline{6-9} 
\textbf{little}   &  &    &  &  & \textbf{work}      & \checkmark  &    &  \\ \cline{1-4} \cline{6-9} 
\textbf{lol}      &  &    &  &  & \textbf{world}     & \checkmark  &    & \checkmark  \\ \cline{1-4} \cline{6-9} 
\textbf{look}     &  & \checkmark    & \checkmark  &  & \textbf{would}     & \checkmark  & \checkmark    & \checkmark  \\ \cline{1-4} \cline{6-9} 
\textbf{love}     & \checkmark  &    & \checkmark  &  & \textbf{year}      &  &    &  \\ \cline{1-4} \cline{6-9} 
\textbf{made}     & \checkmark  &    & \checkmark  &  & \textbf{you}       & \checkmark  & \checkmark    & \checkmark  \\ \cline{1-4} \cline{6-9} 
\textbf{make}     & \checkmark  & \checkmark    & \checkmark  &  & \textbf{your}      & \checkmark  & \checkmark    & \checkmark  \\ \cline{1-4} \cline{6-9} 
\textbf{man}      &  &    &  &  & \textbf{youre}     & \checkmark  & \checkmark    & \checkmark  \\ \cline{1-4} \cline{6-9} 
\end{longtable*}

\newpage
\subsection{Comparison with null models}

\begin{figure*}[ht!]
    \centering
    \includegraphics[width=\textwidth]{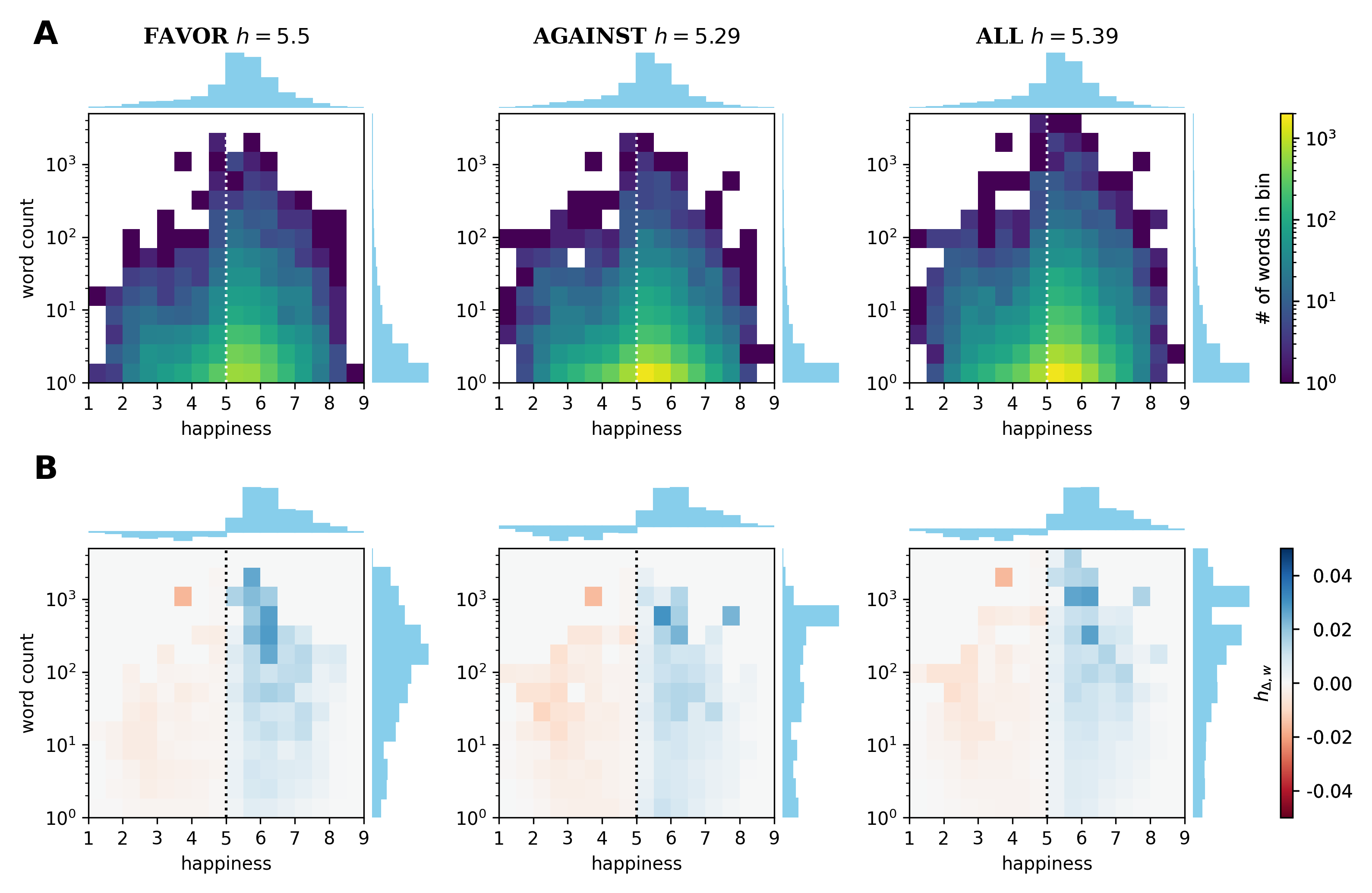}
    \caption{(a) The 2D histogram for word count vs. happiness score, with the corresponding marginal distributions shown (note that each word has a weight of 1) for the \textbf{configuration model}. (b) A 2D histogram of the contributions of words in word count-happiness space to the deviation from neutrality, $h_{\Delta, w} = (h_w - 5) * N_w / \sum_{w^{\prime}}{N_{w^{\prime}}}$, where $h_w$ is the word's happiness score of word and $N_w$ is the number of times the word appears in the corpus. The marginal distributions are also included. Vertical lines at $h=5$ are added to guide the eye.
    }
    \label{fig:hclinton_scorecontrib_count_dist_v2_configmodel}
\end{figure*}

\begin{figure*}[ht!]
    \centering
    \includegraphics[width=\textwidth]{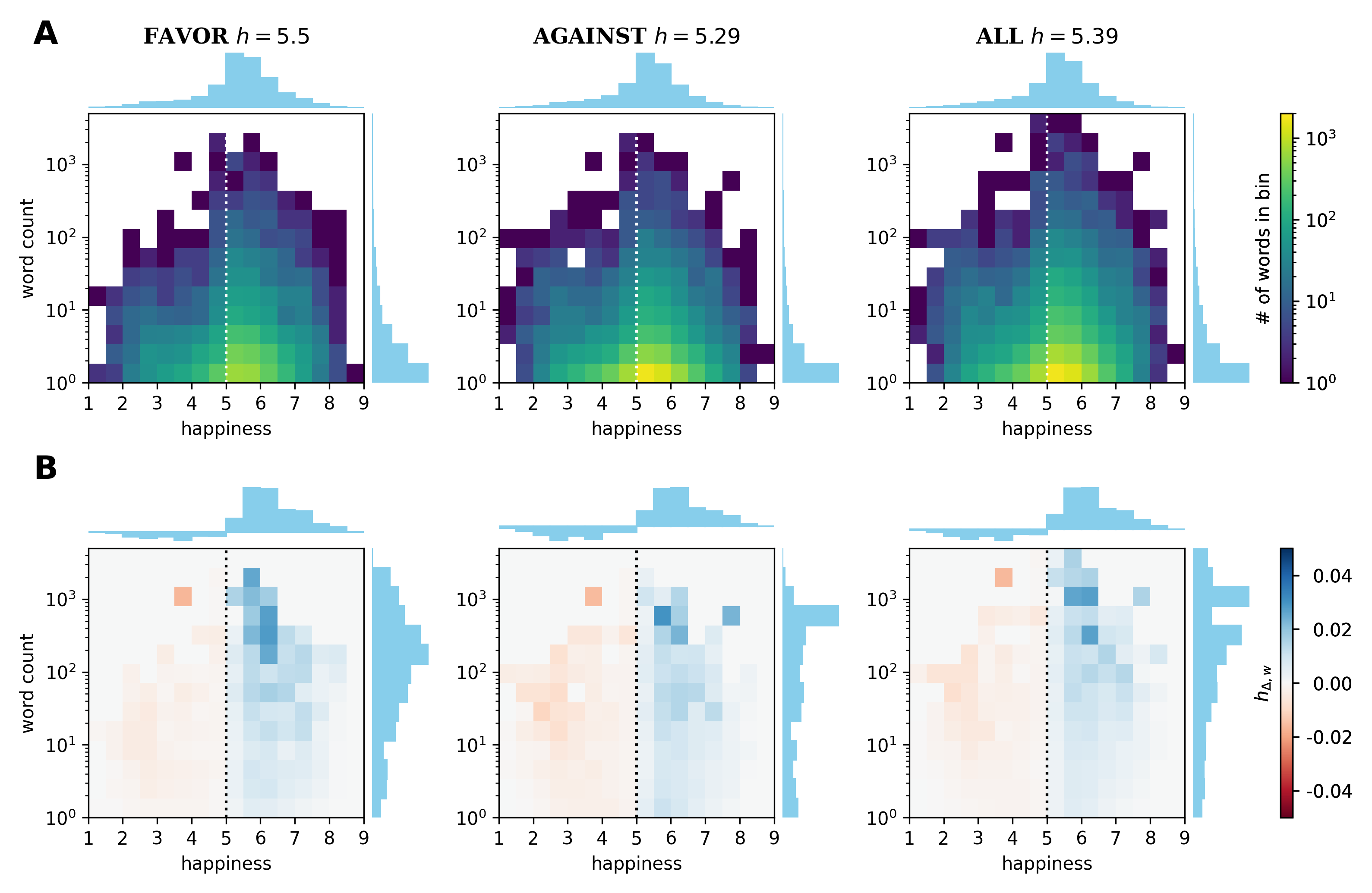}
    \caption{(a) The 2D histogram for word count vs. happiness score, with the corresponding marginal distributions shown (note that each word has a weight of 1) for the \textbf{Erdos-Renyi null model}. (b) A 2D histogram of the contributions of words in word count-happiness space to the deviation from neutrality, $h_{\Delta, w} = (h_w - 5) * N_w / \sum_{w^{\prime}}{N_{w^{\prime}}}$, where $h_w$ is the word's happiness score of word and $N_w$ is the number of times the word appears in the corpus. The marginal distributions are also included. Vertical lines at $h=5$ are added to guide the eye.
    }
    \label{fig:hclinton_scorecontrib_count_dist_v2_ERauto}
\end{figure*}

\begin{figure*}[ht!]
    \centering
    \includegraphics[width=\textwidth]{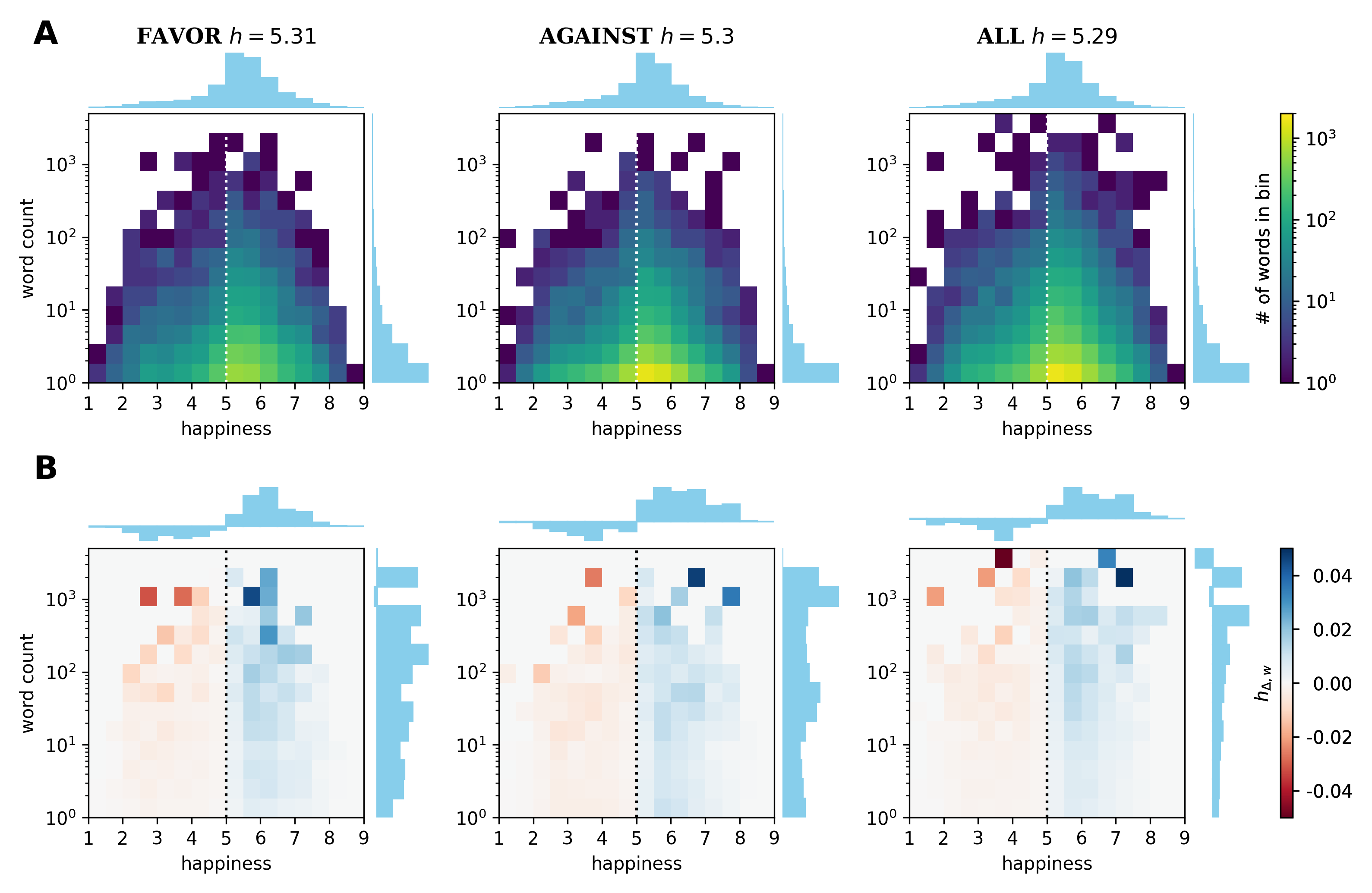}
    \caption{(a) The 2D histogram for word count vs. happiness score, with the corresponding marginal distributions shown (note that each word has a weight of 1) for the \textbf{shuffled score model}. (b) A 2D histogram of the contributions of words in word count-happiness space to the deviation from neutrality, $h_{\Delta, w} = (h_w - 5) * N_w / \sum_{w^{\prime}}{N_{w^{\prime}}}$, where $h_w$ is the word's happiness score of word and $N_w$ is the number of times the word appears in the corpus. The marginal distributions are also included. Vertical lines at $h=5$ are added to guide the eye.
    }
    \label{fig:hclinton_scorecontrib_count_dist_v2_internalshuffle}
\end{figure*}

\begin{figure*}[ht!]
    \centering
    \includegraphics[width=\textwidth]{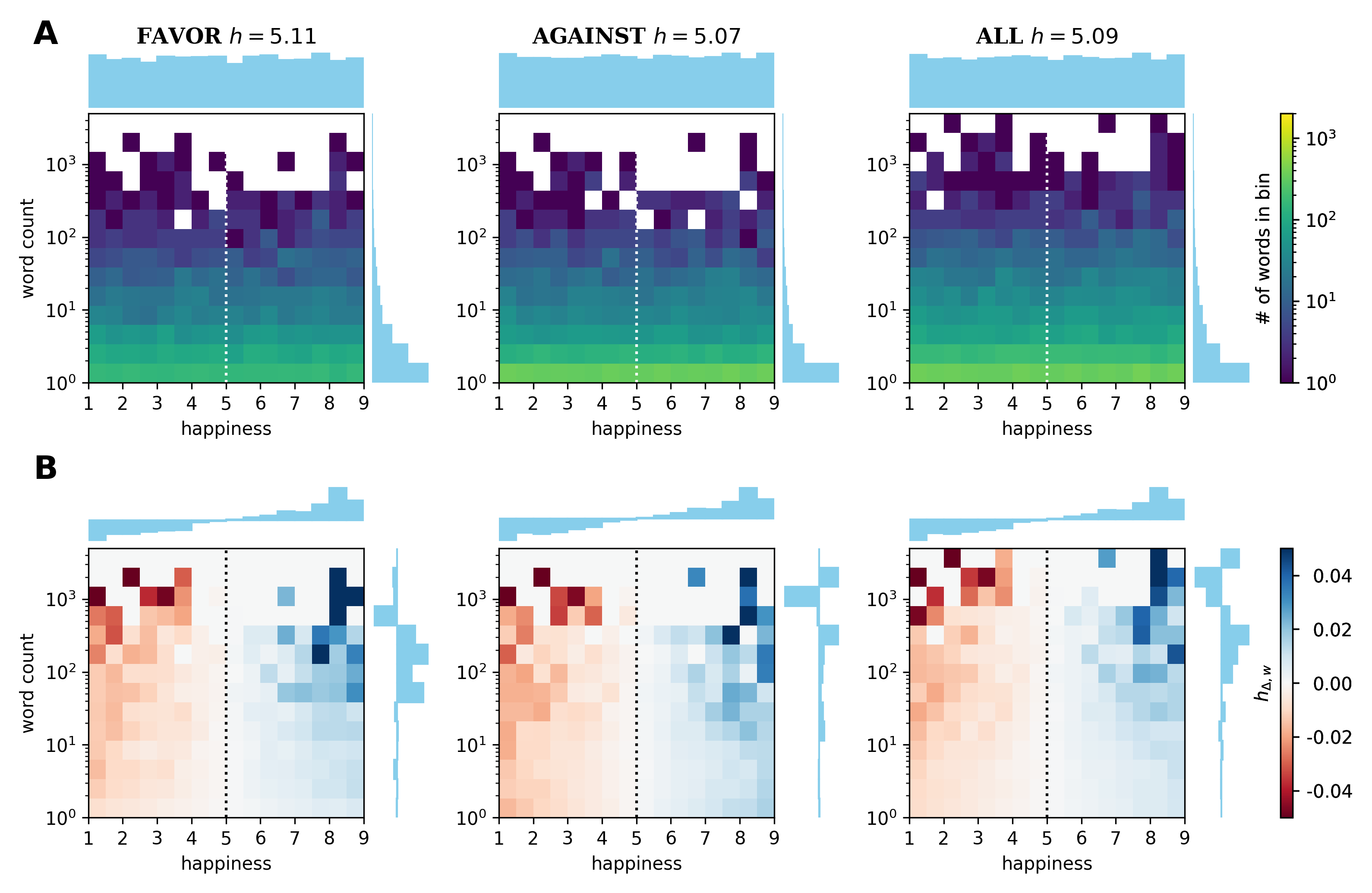}
    \caption{(a) The 2D histogram for word count vs. happiness score, with the corresponding marginal distributions shown (note that each word has a weight of 1) for the \textbf{uniform score model}. (b) A 2D histogram of the contributions of words in word count-happiness space to the deviation from neutrality, $h_{\Delta, w} = (h_w - 5) * N_w / \sum_{w^{\prime}}{N_{w^{\prime}}}$, where $h_w$ is the word's happiness score of word and $N_w$ is the number of times the word appears in the corpus. The marginal distributions are also included. Vertical lines at $h=5$ are added to guide the eye.
    }
    \label{fig:hclinton_scorecontrib_count_dist_v2_uniform}
\end{figure*}

\begin{figure*}[ht!]
    \centering
    \includegraphics[width=\textwidth]{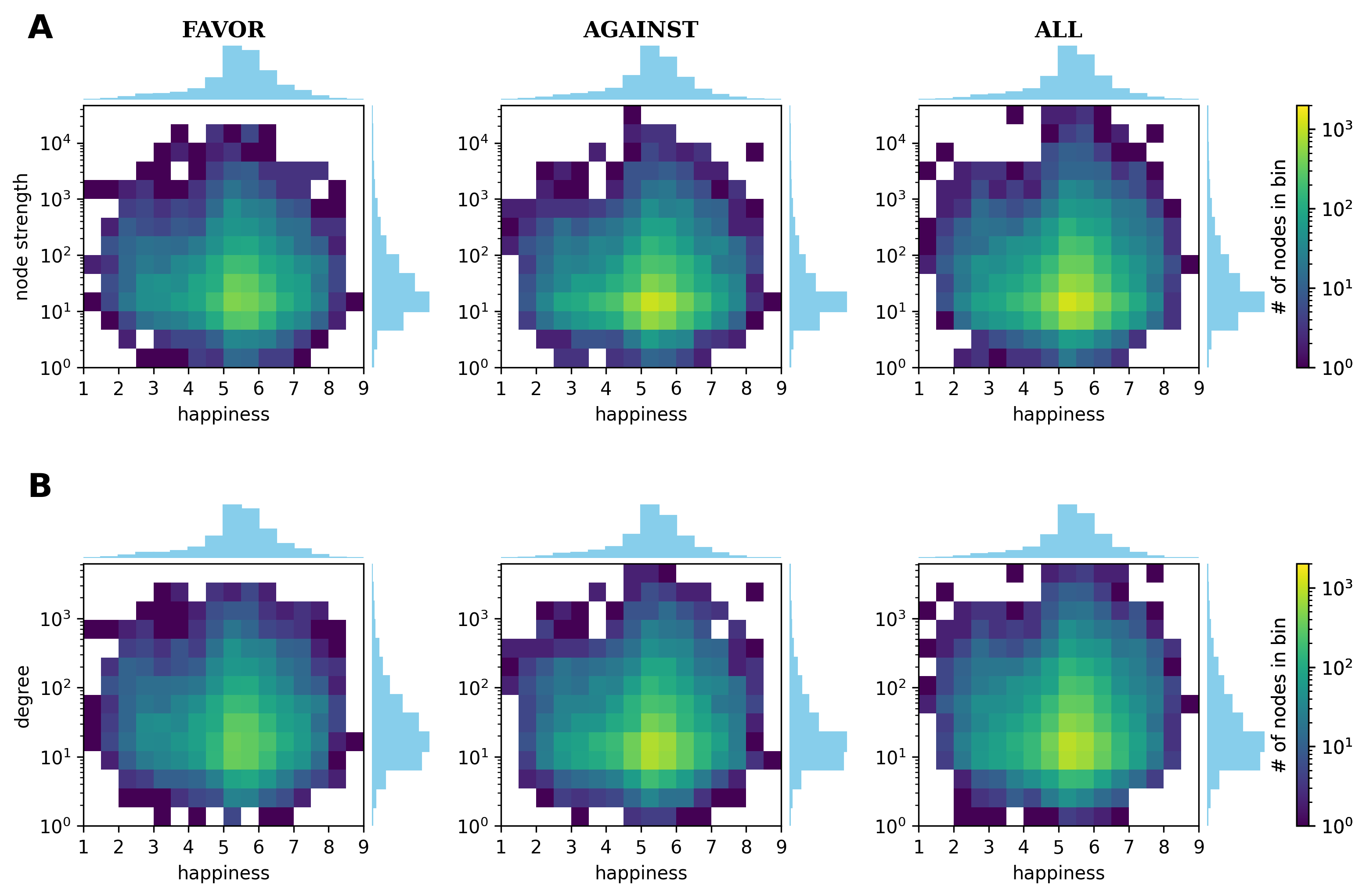}
    \caption{2D histograms for both the node strengths (top row) and degree (bottom row) vs. happiness scores for the \textbf{configuration model}.}
    \label{fig:hclinton_score_degree_dist_v2_configmodel}
\end{figure*}

\begin{figure*}[ht!]
    \centering
    \includegraphics[width=\textwidth]{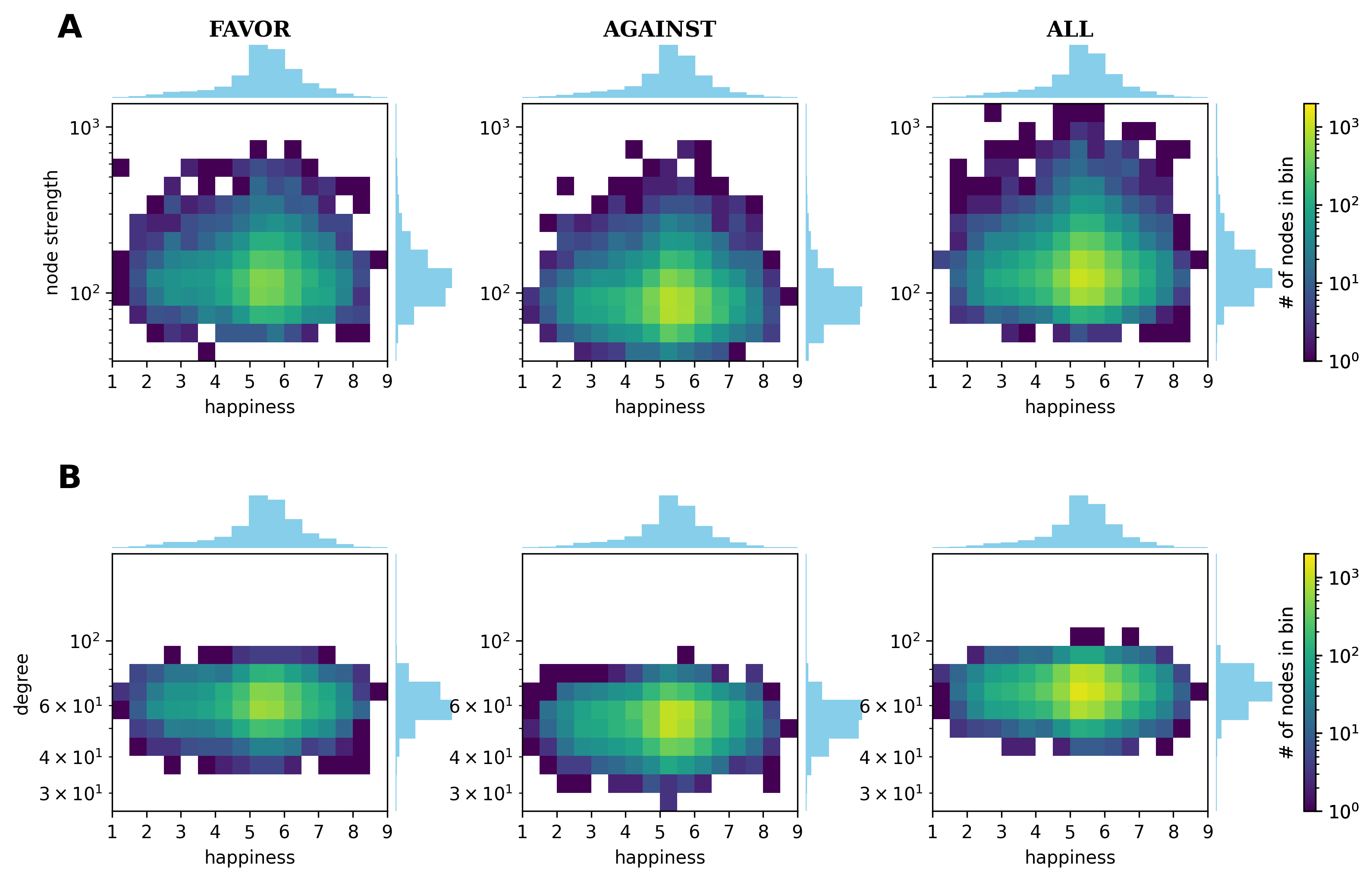}
    \caption{2D histograms for both the node strengths (top row) and degree (bottom row) vs. happiness scores for the \textbf{Erdos-Renyi null model}.}
    \label{fig:hclinton_score_degree_dist_v2_ERauto}
\end{figure*}

\begin{figure*}[ht!]
    \centering
    \includegraphics[width=\textwidth]{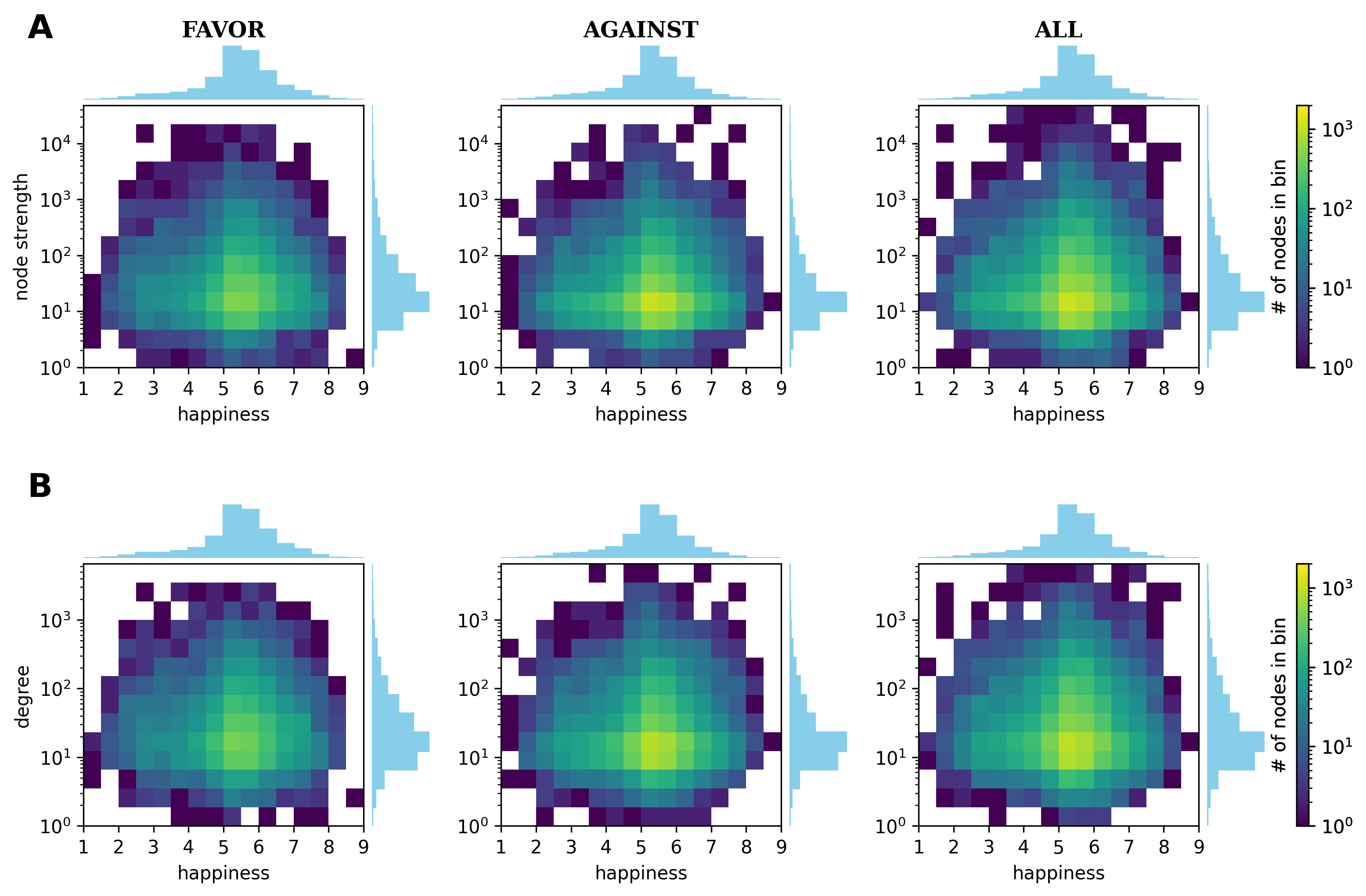}
    \caption{2D histograms for both the node strengths (top row) and degree (bottom row) vs. happiness scores for the \textbf{shuffled score model}.}
    \label{fig:hclinton_score_degree_dist_v2_internalshuffle}
\end{figure*}

\begin{figure*}[ht!]
    \centering
    \includegraphics[width=\textwidth]{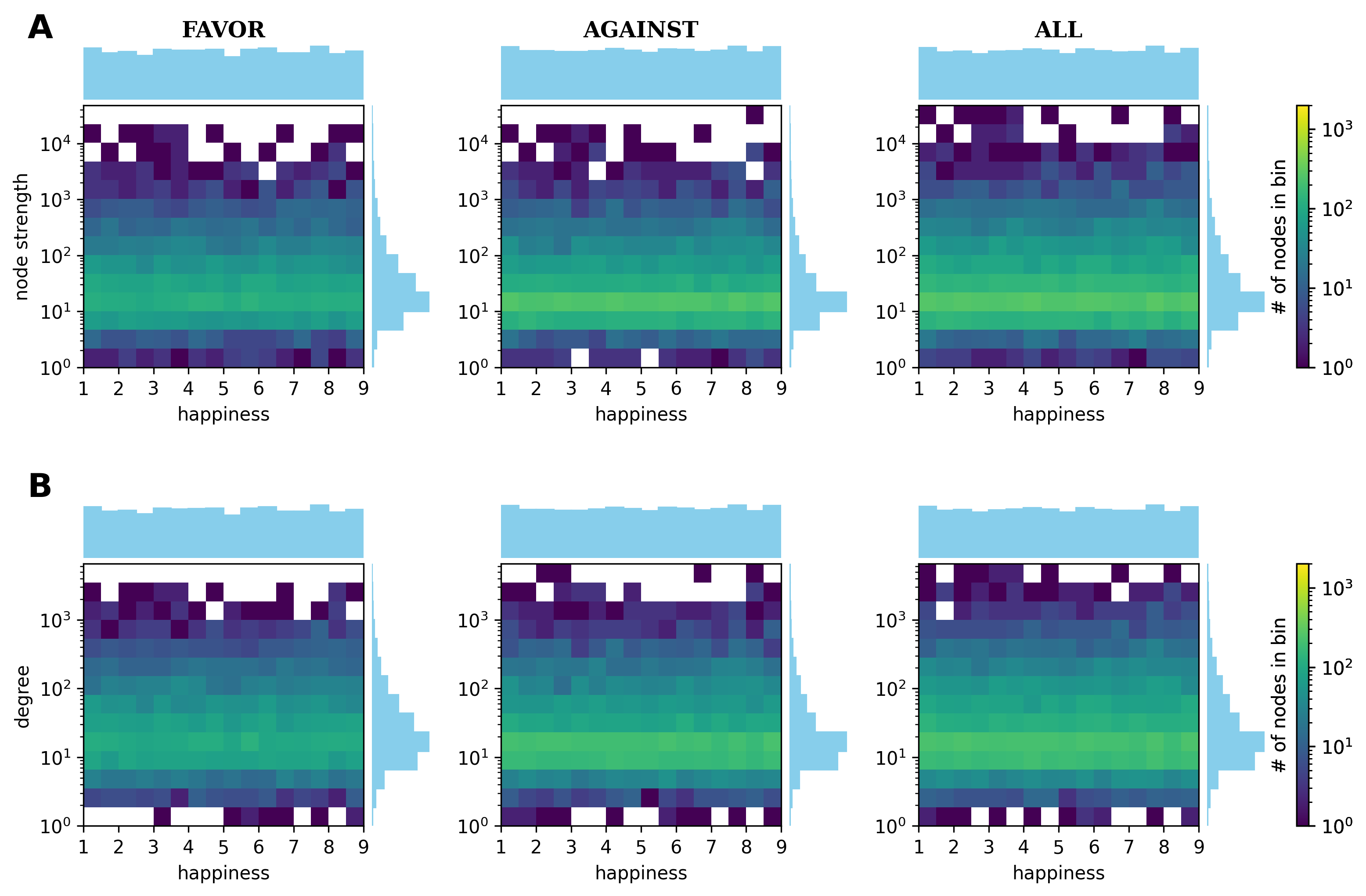}
    \caption{2D histograms for both the node strengths (top row) and degree (bottom row) vs. happiness scores for the \textbf{uniform score model}.}
    \label{fig:hclinton_score_degree_dist_v2_uniform}
\end{figure*}

\begin{figure*}[ht!]
    \centering
    \includegraphics[width=\textwidth]{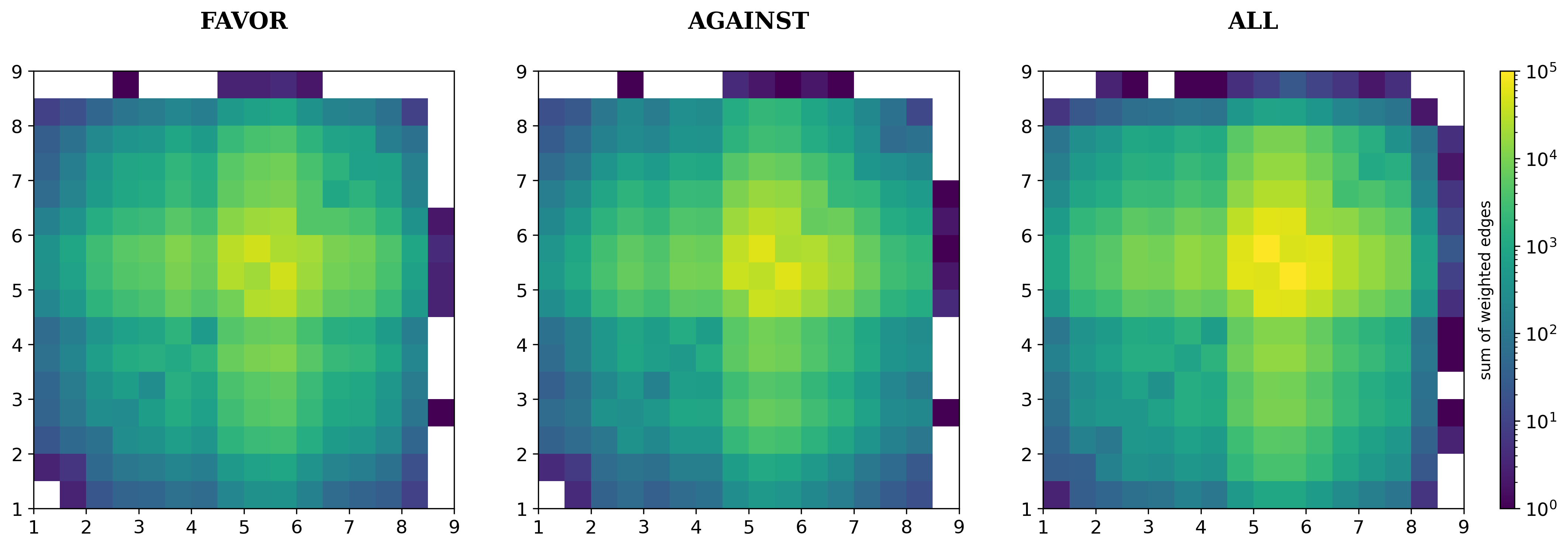}
    \caption{The happiness scores of each pair of nodes for the \textbf{configuration model} are plotted in this 2D histogram. Each pair of nodes is weighted by the weight of the edge connecting them. We made the histogram to be symmetric about the 45$\degree$ line so that one can analyze it from either the horizontal or vertical direction.}
    \label{fig:hclinton_scorepair_happiness_heatmap_v2_configmodel}
\end{figure*}

\begin{figure*}[ht!]
    \centering
    \includegraphics[width=\textwidth]{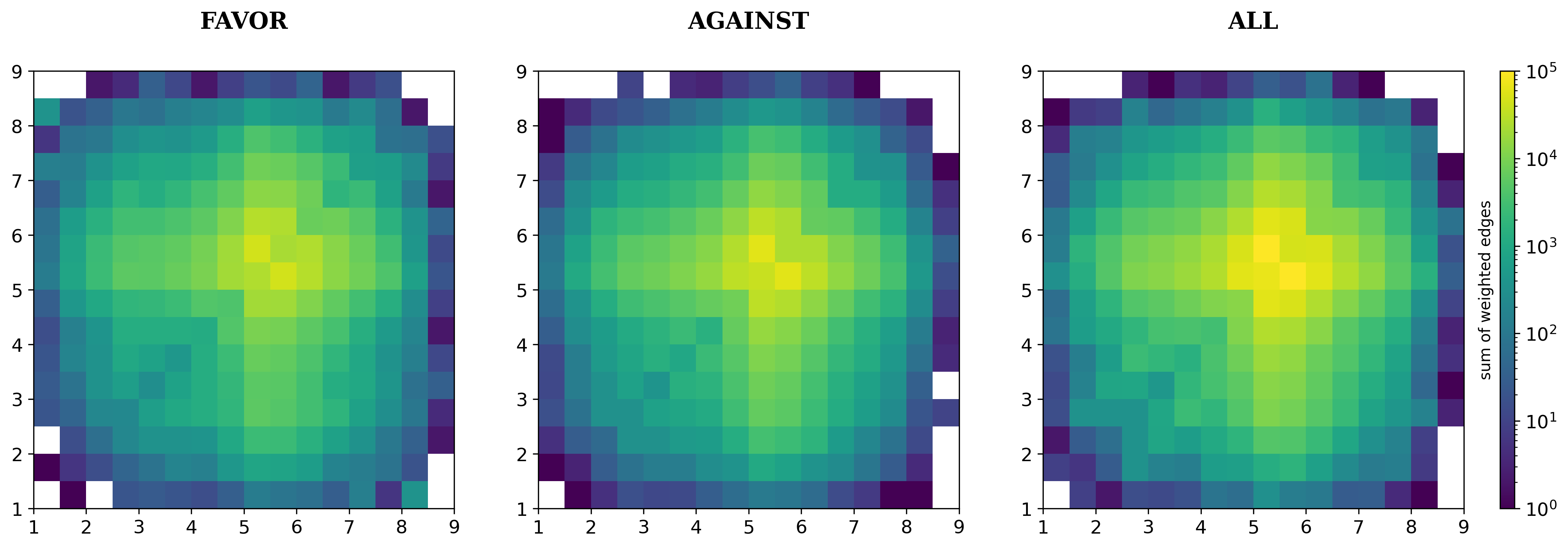}
    \caption{The happiness scores of each pair of nodes for the \textbf{Erdos-Renyi null model} are plotted in this 2D histogram. Each pair of nodes is weighted by the weight of the edge connecting them. We made the histogram to be symmetric about the 45$\degree$ line so that one can analyze it from either the horizontal or vertical direction.}
    \label{fig:hclinton_scorepair_happiness_heatmap_v2_ERauto}
\end{figure*}

\begin{figure*}[ht!]
    \centering
    \includegraphics[width=\textwidth]{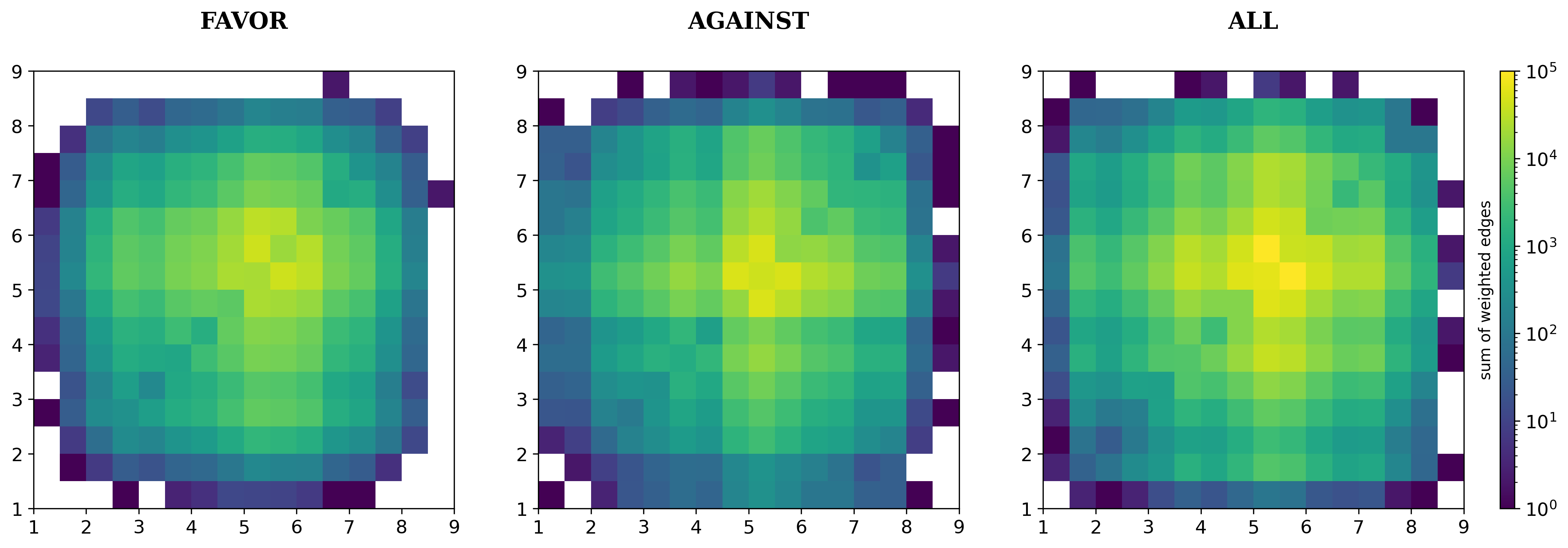}
    \caption{The happiness scores of each pair of nodes for the \textbf{shuffled score model} are plotted in this 2D histogram. Each pair of nodes is weighted by the weight of the edge connecting them. We made the histogram to be symmetric about the 45$\degree$ line so that one can analyze it from either the horizontal or vertical direction.}
    \label{fig:hclinton_scorepair_happiness_heatmap_v2_internalshuffle}
\end{figure*}

\begin{figure*}[ht!]
    \centering
    \includegraphics[width=\textwidth]{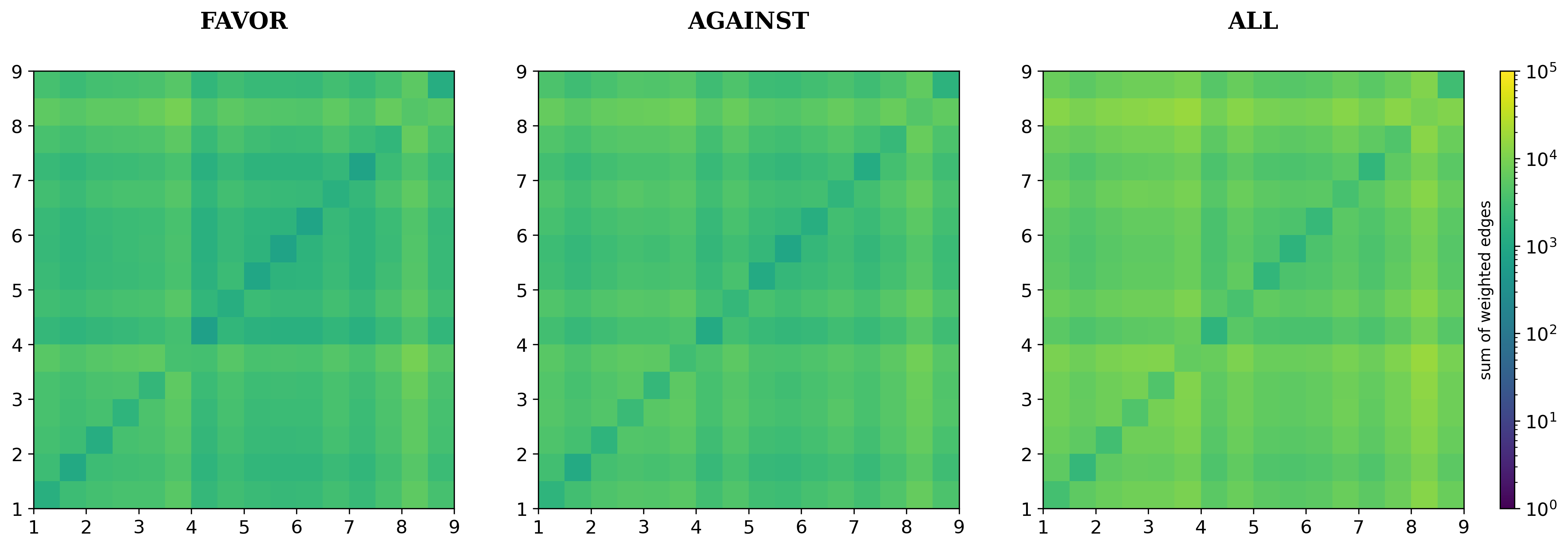}
    \caption{The happiness scores of each pair of nodes for the \textbf{uniform score model} are plotted in this 2D histogram. Each pair of nodes is weighted by the weight of the edge connecting them. We made the histogram to be symmetric about the 45$\degree$ line so that one can analyze it from either the horizontal or vertical direction.}
    \label{fig:hclinton_scorepair_happiness_heatmap_v2_uniform}
\end{figure*}

We include here the histograms (Figures~\ref{fig:hclinton_scorecontrib_count_dist_v2_configmodel}-\ref{fig:hclinton_scorepair_happiness_heatmap_v2_uniform}) corresponding to Figures~\ref{fig:hclinton_scorecontrib_count_dist_v2}-\ref{fig:hclinton_scorepair_happiness_heatmap_v2} for the null models using the configuration network model, the Erdos-Renyi model, the shuffled score model, and the uniform score model (Section~\ref{subsec:method-network-characterization}).

\FloatBarrier

\subsection{Disparity filter}

\begin{figure*}[ht!]
    \centering
    \includegraphics[width=\textwidth]{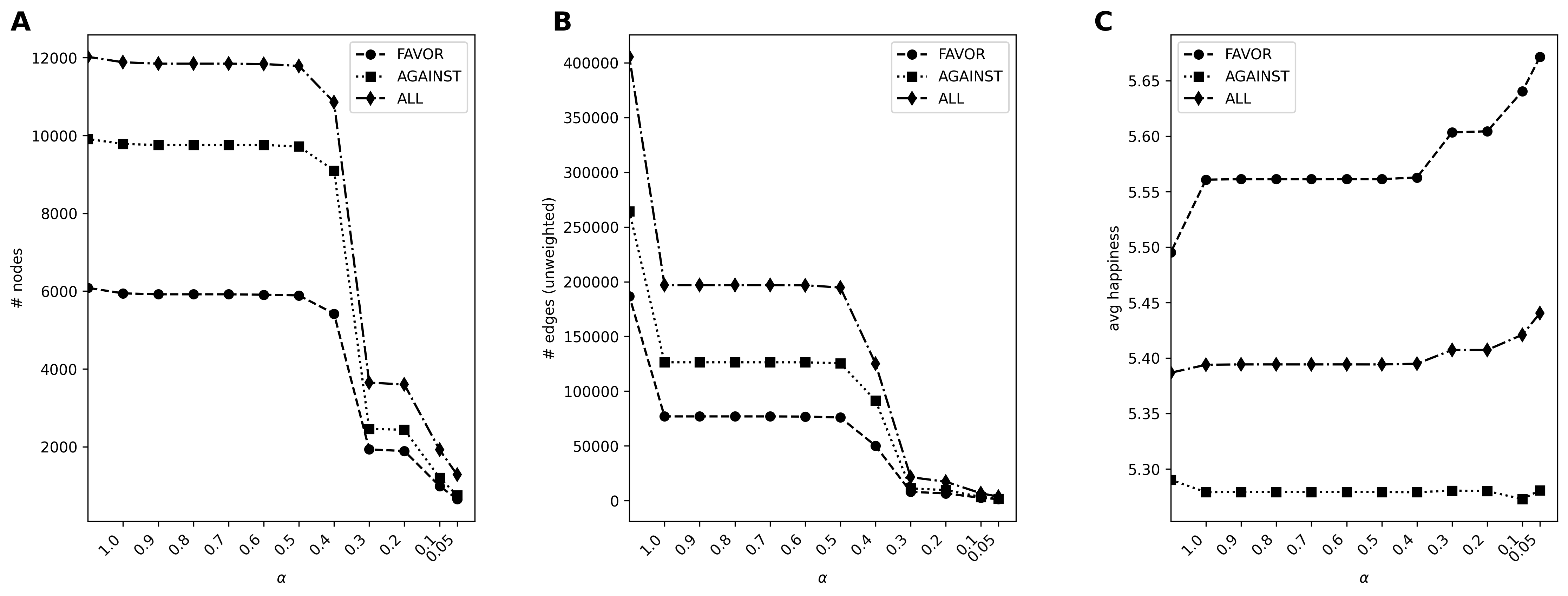}
    \caption{Graph (a) order, (b) size and (c) average happiness score weighted by word count as a function of the threshold used in the disparity filter. $\alpha=1$ indicates that the top words from Twitter were removed, but the disparity filter was not applied, while the points on the y-axis correspond to the values from the complete network without any backboning.}
    \label{fig:hclinton_graphprop_by_threshold}
\end{figure*}

\begin{figure*}[ht!]
    \centering
    \includegraphics[width=0.8\textwidth]{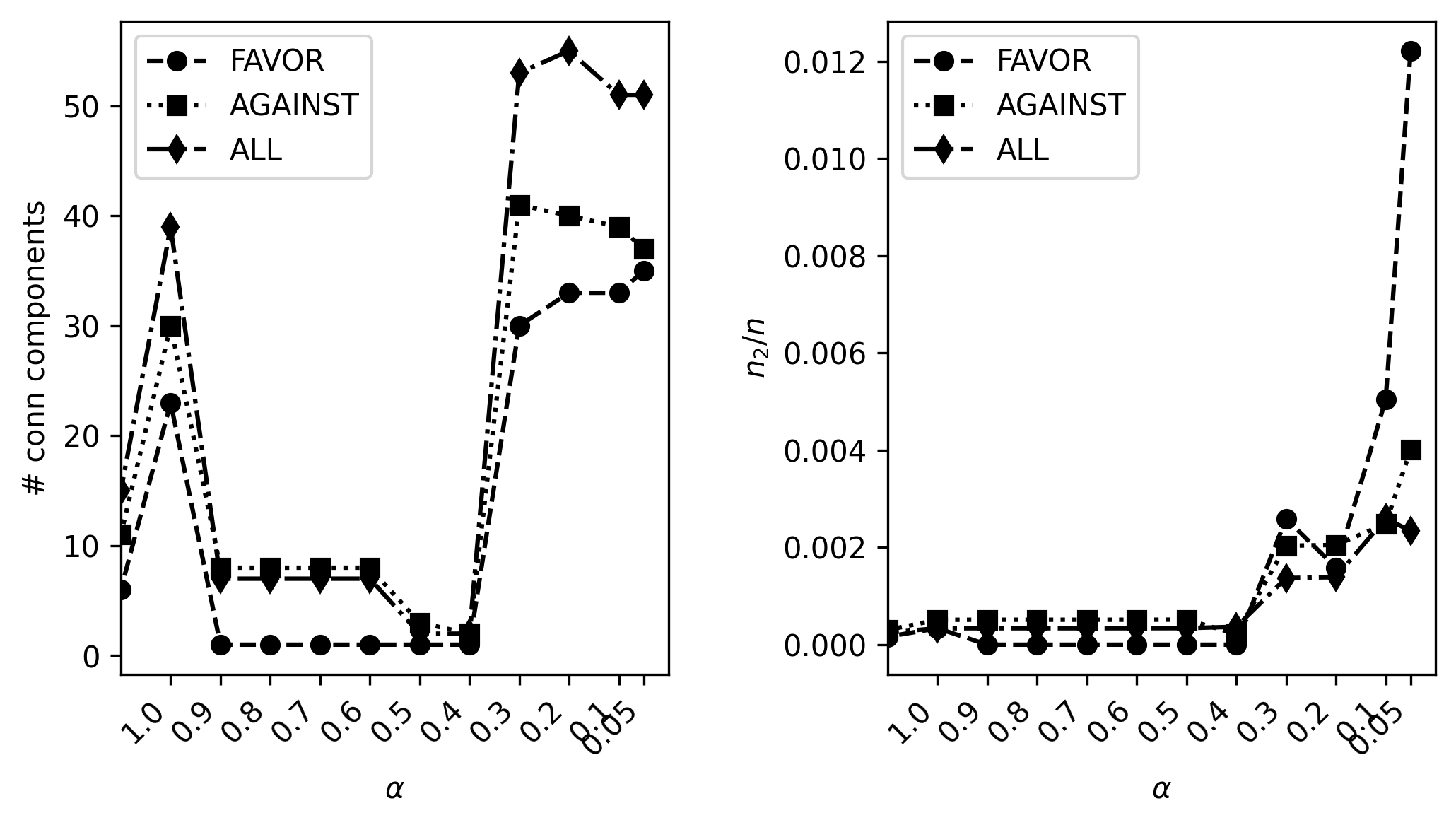}
    \caption{Number of connected components and relative size of the second-largest component to the largest component for all thresholds.}
    \label{fig:hclinton_components_by_threshold}
\end{figure*}

Figure~\ref{fig:hclinton_graphprop_by_threshold} shows how the size and order of the backbone vary depending on the significance level $\alpha$ chosen after removing the most commonly used words and applying the disparity filter, with $\alpha=1$ corresponding to the network without the most common words on Twitter and the values on the y-axis corresponding to the complete network from which neither the most common words on Twitter nor any likely spurious edges were removed. The removal of words from Twitter eliminated several edges from the network, but few edges are removed by the disparity filter until the threshold value reaches $\alpha=0.4$. As the value of $\alpha$ decreases, more edges are eliminated, but this also means more nodes are likely to be removed from the backbone. As a result, the number of components does not change monotonically with $\alpha$ (Figure~\ref{fig:hclinton_components_by_threshold}). Note that since the disparity filter removes isolated nodes, there is a huge drop in the number of components from $\alpha=1$ to $\alpha=0.9$ even if the number of nodes does not change much. The giant component is also by far the largest of all the components, as can be deduced in the plot of the fraction $n_2 / n$, where $n$ is the number of nodes in the entire network and $n_2$ is the number of nodes in the second largest component  (Figure~\ref{fig:hclinton_components_by_threshold}).

\begin{figure*}[ht!]
    \centering
    \includegraphics[width=\textwidth]{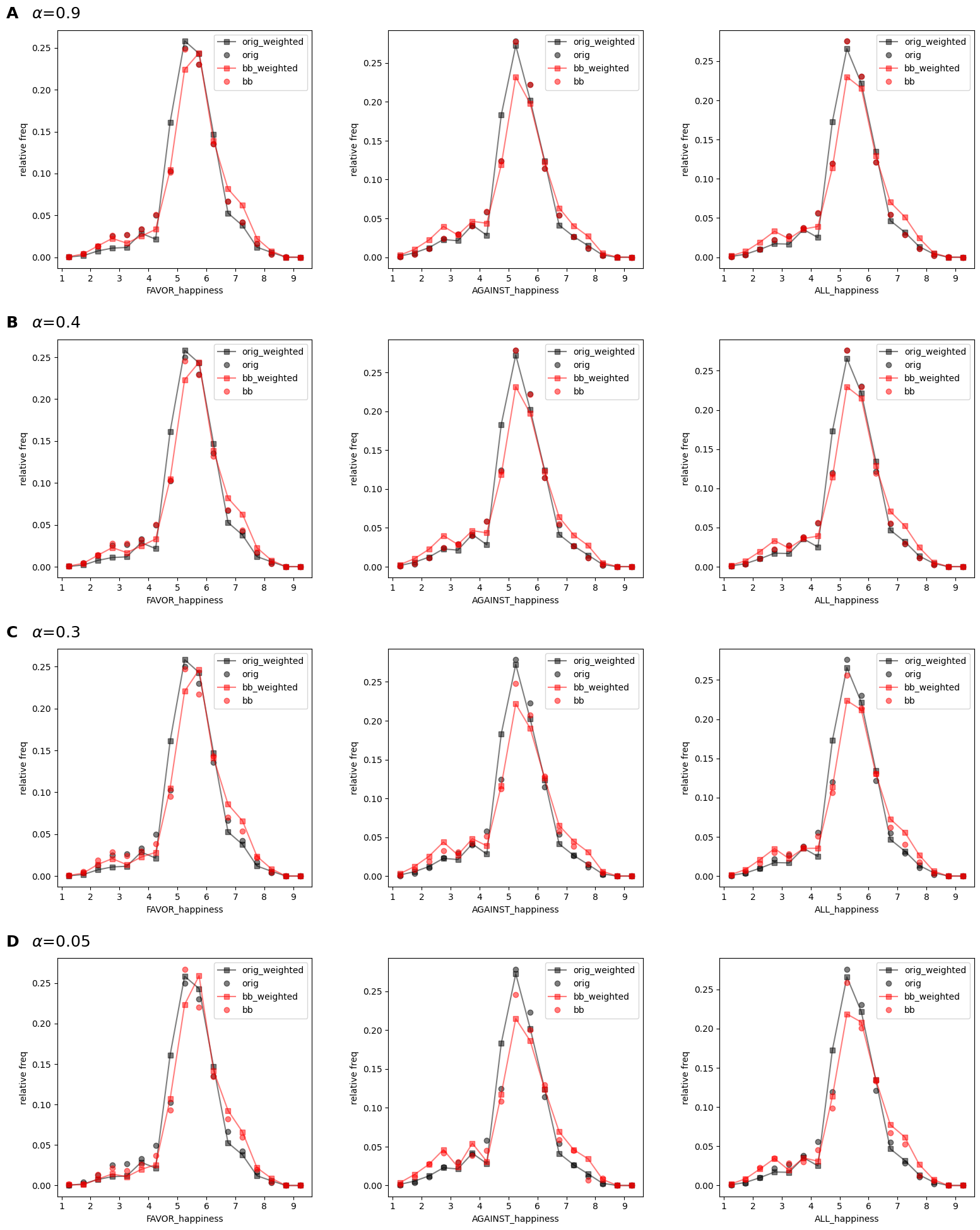}
    \caption{Relative frequencies of happiness scores in the backbone for significance levels $\alpha=0.9, 0.4, 0.3, 0.05$. Curves marked ``orig'' refer to the original network, while those marked ``bb'' refer to the backbone. The suffix ``\_weighted'' indicates that the words are weighted by their word counts.}
    \label{fig:hclinton_scoredist_by_threshold}
\end{figure*}

The resulting score distributions, however, are very similar regardless of the threshold (Figure~\ref{fig:hclinton_scoredist_by_threshold}). As the most commonly used words on Twitter are also mostly neutral, removing these manually from the network increases the relative frequency of positive and negative words compared to the original network. Note that this change is due to the removal of the most commonly used words on Twitter; the disparity filter itself only minimally alters the score distribution except for low values of $\alpha$, where a slight decrease in the relative frequency of negative words in the ``favor'' corpus is observed. This implies that in terms of the criteria set by the disparity filter, the relevance of nodes does not depend much on their happiness scores, as both neutral and non-neutral words get filtered out.

\begin{figure*}[ht!]
    \centering
    \includegraphics[width=\textwidth]{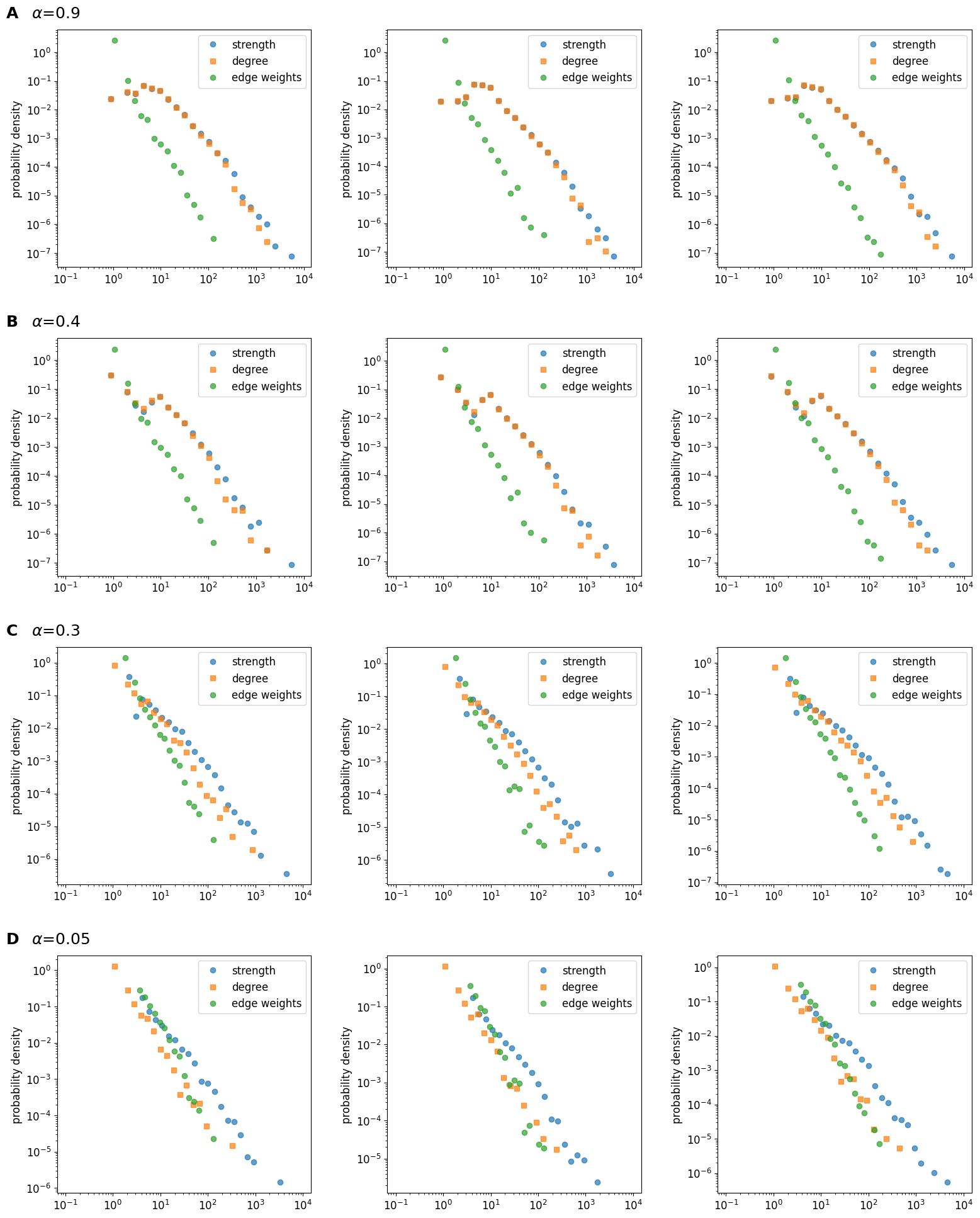}
    \caption{The distributions of the node degrees, node strengths, and edge weights for significance levels $\alpha=0.9, 0.4, 0.3, 0.05$.}
    \label{fig:hclinton_edgeweight_degree_dist_by_threshold}
\end{figure*}

As the disparity filter becomes more restrictive, it removes more edges and alters the degree distribution (Figure~\ref{fig:hclinton_edgeweight_degree_dist_by_threshold}), reducing the peak from around $k=10$ to around $k=1$. The filter also disproportionately targets edges with lower weights. Whereas the maximum degrees and edge weights remain similar across different values of $\alpha$, edges with weights close to 1 have been completely eliminated with $\alpha=0.05$.

\begin{figure*}[ht!]
    \centering
    \includegraphics[width=\textwidth]{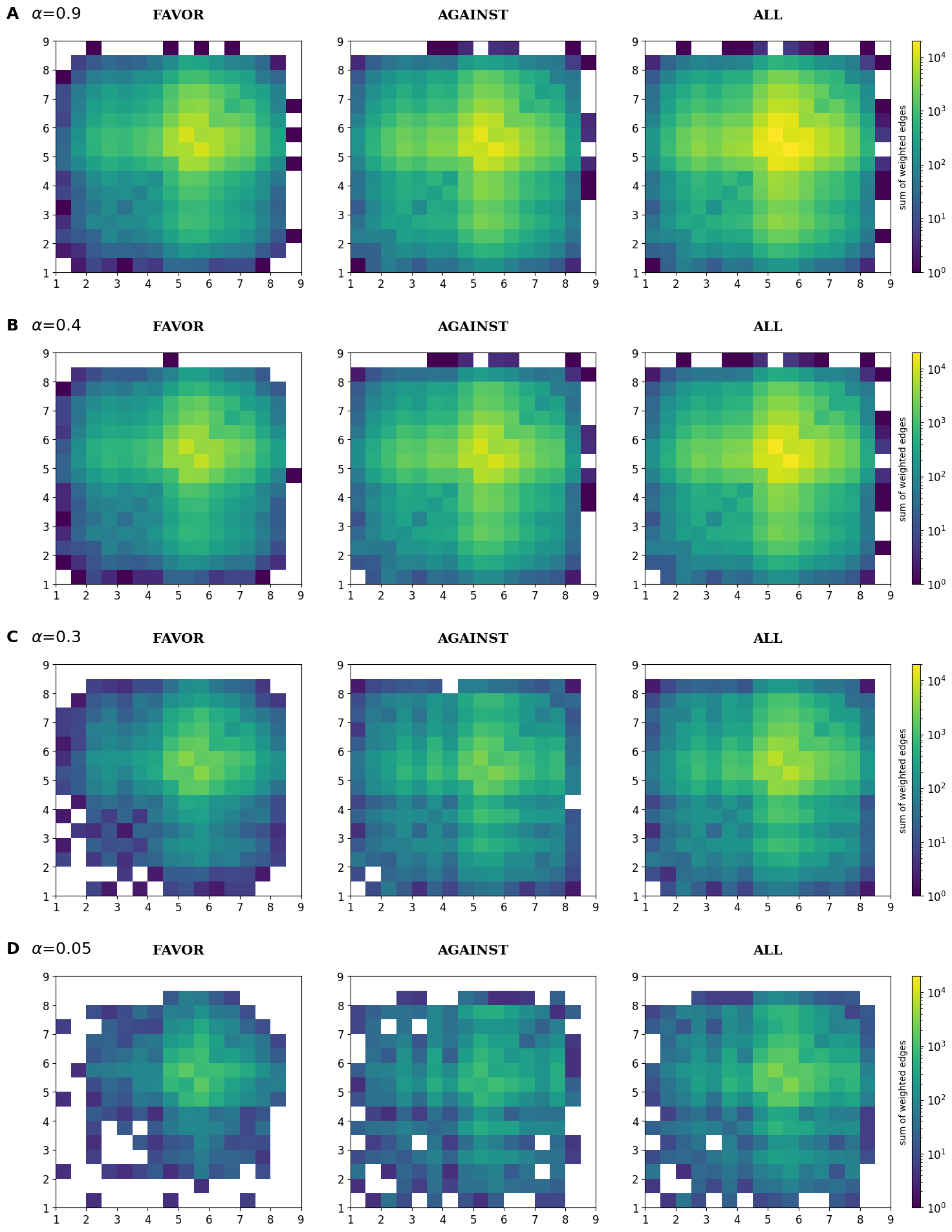}
    \caption{2D histogram of scores of connected nodes weighted by the edge weights for significance levels $\alpha=0.9, 0.4, 0.3, 0.05$. Note that the heat map is made to be symmetric about the 45$\degree$ line so it can be analyzed from either the horizontal or vertical direction.}
    \label{fig:hclinton_scorepair_by_threshold}
\end{figure*}


\end{document}